\pdfoutput=1

\documentclass[11pt]{article}

\usepackage{EMNLP2022}
\usepackage{lipsum}
\usepackage{color, colortbl}
	
\definecolor{Gray}{gray}{0.9}

\usepackage{times}
\usepackage{latexsym}
\usepackage{amsmath, amsfonts}
\usepackage{tabularx,booktabs}
\usepackage{adjustbox}

\usepackage[T1]{fontenc}

\usepackage[utf8]{inputenc}

\usepackage{microtype}

\usepackage{inconsolata}
\usepackage{xcolor}
\usepackage[ruled]{algorithm2e}

\SetCommentSty{mycommfont}

\SetKwInput{KwInput}{Input}                
\SetKwInput{KwOutput}{Output}              

\title{Meta-learning Pathologies from Radiology Reports using Variance Aware Prototypical Networks}

\author{Arijit Sehanobish \hspace{.1cm}  Kawshik Kannan\thanks{~~Equal Contribution} \hspace{.1cm} Nabila Abraham \hspace{.1cm} Anasuya Das \hspace{.1cm}  Benjamin Odry \\
  Covera Health \\
  New York City, NY \\
 \texttt{\{arijit.sehanobish, kawshik.kannan, nabila.abraham,}  \\
 \texttt{anasuya.das, benjamin.odry\}@coverahealth.com} \\
}
\begin{document}
\maketitle
\begin{abstract}
Large pretrained Transformer-based language models like BERT and GPT have changed the landscape of Natural Language Processing (NLP). However, fine tuning such models still requires a large number of training examples for each target task, thus annotating multiple datasets and training these models on various downstream tasks becomes time consuming and expensive. In this work, we propose a simple extension of the Prototypical Networks for few-shot text classification. Our main idea is to replace the class prototypes by Gaussians and introduce a regularization term that encourages the examples to be clustered near the appropriate class centroids. Experimental results show that our method outperforms various strong baselines on $13$ public and $4$ internal datasets. Furthermore, we use the class distributions as a tool for detecting potential out-of-distribution (OOD) data points during deployment.
\end{abstract}

\section{Introduction}
Pretrained Transformer-based language models (PLMs) have achieved great success on many NLP tasks~\citep{devlin-etal-2019-bert, NEURIPS2020_1457c0d6}, but still need a large number of in-domain labeled examples for finetuning~\citep{finetuning}. Learning to learn~\citep{learningtolearn, schmidhuber:1987:srl, Bengio97onthe} from limited supervision is an important problem with widespread application in areas where obtaining labeled data can be difficult or expensive. To that end, meta-learning methods have been proposed as effective solutions for few-shot learning~\citep{metalearning_survey}. Current applications of such meta-learning methods have shown improved performance in few-shot learning for vision tasks such as learning to classify new image classes within a similar dataset. Namely, on classical few-shot image classification benchmarks, the training tasks are sampled from a “single” larger dataset (for ex: Omniglot~\citep{omniglot} and miniImageNet~\citep{matchingnet}), and the label space contains the same task structure for all tasks. There has been a similar trend of such classical methods in NLP as well~\citep{fewshottext}. In contrast, in text classification tasks, the set of source tasks available during training and target tasks during evaluation can range from sentiment analysis to grammatical acceptability judgment~\citep{bansal-etal-2020-learning, bansal2020self}. In recent works~\citep{wang2021gradtask}, the authors use a range of different source tasks (different not only in terms of input domain, but also their task structure i.e. label semantics, and number of labels) for meta-training and show successful performance on a wide range of downstream tasks. In spite of this success, meta-training on various source tasks is quite challenging as it requires resistance to overfitting to certain source tasks due to its few-shot nature and more task-specific adaptation due to the distinct nature among tasks~\citep{metaoverfitting}. 

However, in medical NLP, collecting large number of diverse labeled datasets is difficult. In our institution, we collect high quality labeled radiology reports (which are always used as held out test data) and use it to train our internal annotators who then annotate our unlabeled data. This training process is expensive and time consuming. Our annotation process is described in section~\ref{sec:annot}. Thus a natural question is: if we have a large labeled dataset consisting of a lot of classes, can we use it to meta-train a model that can be used on a large number of downstream datasets where we have little to no training examples? This is a challenging problem as the reports can be structured differently based on the report type and there can be a substantial variation in writing style across radiologists from different institutions. Our main goal is to build out a set of extensible pipelines that can generalize to new pathologies typically in new sub-specialties while also generalizing across different health systems. In addition, the exact definition of the pathologies and their severity change can change depending on the clinical use case. This makes fully supervised approaches that rely on large labeled datasets expensive. Having few-shot capabilities allows us to annotate a handful of cases and rapidly expand the list of pathologies we can detect and classify. In addition, we can use our approach to generate pseudo labels for rare pathologies and enrich our validation and test sets for annotation by an in-house clinical team. Lastly our approach can be extended to support patient search and define custom cohorts of patients.

Our contributions in this work are the following: \textbf{(1)} We develop a novel loss function that extends the vanilla prototypical networks and introduce a regularization term that encourages tight clustering of examples near the class prototypes. \textbf{(2)} We meta-train our models on a large labeled dataset on shoulder MRI reports (single domain) and show good performance on $4$ diverse downstream classification tasks on radiology reports on knee, cervical spine and chest. In addition to our internal datasets, we show superior performance of our method on $13$ public benchmarks over well-known methods like Leopard. Our model is very simple to train, easy to deploy unlike gradient based methods and just requires a few additional lines of codes to a vanilla prototypical network trainer. \textbf{(3) } We deploy our system and use the dataset statistics to inform out-of-distribution (OOD) cases.
\section{Related Work}
There are three common approaches to meta-learning: metric-based, model-based, and optimization-based. Model agnostic meta-learning (MAML)~\citep{pmlr-v70-finn17a} is an optimization-based approach to meta-learning which is agnostic to the model architecture and task specification. Over the years, several variants of the method have shown that it is an ideal candidate for learning to learn from diverse tasks~\citep{reptile, anil, bansal2020self}. However, to solve a new task, MAML type methods would require training a new classification layer for the task. In contrast, metric-based approaches, such as prototypical networks~\citep{matchingnet, protonets}, being non-parametric in nature can handle varied number of classes and thus can be easily deployed. Given the simple nature of prototypical networks, a lot of work has been done to improve them~\citep{pmlr-v97-allen19b, 9010263, ding2022fewshot, wang2021gradtask}. Prototypical networks usually construct a class prototype (mean) using the support vectors to describe the class and, given a query example, assigns the class whose class prototype is closest to the query vector. In~\citep{pmlr-v97-allen19b}, the authors use a mixture of Gaussians to describe the class conditional distribution and in~\citep{9010263}; the authors try to model an unknown general class distribution. In~\citep{ding2022fewshot}, the authors use spherical Gaussians and a KL-divergence type function between the Gaussians to compute the function $d$ in equation~\ref{eqn:proto}. However, the function used by the above authors is not a true metric, i.e. does not satisfy the triangle inequality. Triangle inequality is implicitly important since we use this metric as a form of distance which we optimize, so it makes sense to use a true metric. In this work we replace it by the Wasserstein distance which is a true metric and add in a regularization term that encourages the $L_2$ norm of the covariance matrices to be small, encouraging the class examples to be clustered close to the centroid. One of our main reasons to work with Gaussians is due to the closed form formula of the Wasserstein distance. 

Few shot learning (FSL) in the medical domain has been mostly focused in computer vision~\citep{SINGH2021108111}. There are only a few works that have applied FSL in medical NLP~\citep{metanlp} but most of those works have only focused on different tasks on MIMIC-III~\citep{Johnson2016} which is a single domain dataset (patients from ICU and one hospital system). To the best of our knowledge, ours is the first study to successfully apply FSL on a diverse set of medical datasets (diverse in terms of tasks and patient populations). 
\section{Datasets}~\label{sec:datasets}
\begin{figure}[h]
\includegraphics[width=\columnwidth]{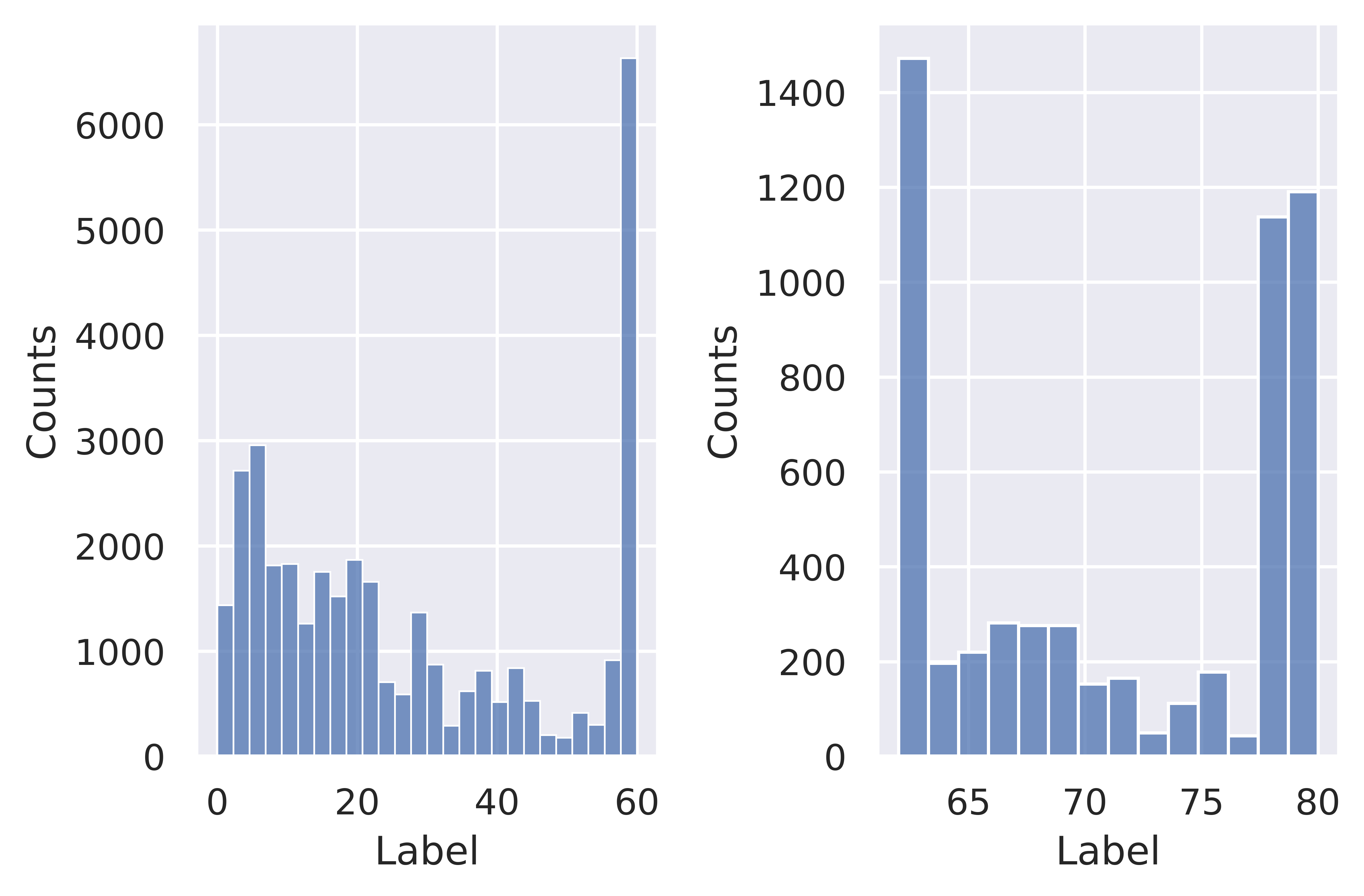}
\caption{Histogram showing the label distribution in (left) train and (right) validation dataset.}
    \label{fig:histplot}
\end{figure}

All our internal datasets are MRI radiology reports detailing various pathologies in different body parts. Our models are meta-trained on a dataset of shoulder pathologies which is collected from 74 unique and de-identified institutions in the United States. 60 labels are chosen for training and 20 novel labels are chosen for validation. The number of training labels is similar to some well-known image datasets~\citep{omniglot, matchingnet, WahCUB_200_2011}. This diverse dataset has a rich label space detailing multiple structures in shoulder, granular pathologies and their severity levels in each structure. The relationship between the granularity/severity of these pathologies at different structures can be leveraged for other pathologies in different body parts and may lead to successful transfer to various downstream tasks. The labels are split such that all pathologies in a given structure appear at either training or validation but not both. More details about the label space can be found in section~\ref{sec:metatrain}.  The figure~\ref{fig:histplot} and the table~\ref{tab:metatrain_stats} shows the distribution of labels and an example of this dataset can be found in figure~\ref{fig:shoulder_example}.  
Our metalearner is applied to $4$ downstream binary classification tasks spanning different sub-specialities (cancer screening, musculoskeletal radiology, and neuro-radiology) that are both common as well as clinically important. The statistics for each task are given in table~\ref{tab:test_datasets} : \textbf{(1)} High risk cancer screening for lung nodules (according to Fleischner guidelines~\citep{Nair806} which bucket patients at high-risk of lung cancer and requiring follow up imaging immediately or within 3 months as belonging to Category High Risk ; we consider patients not at high-risk as Low Risk), \textbf{(2)}  Complete Anterior Cruciate Ligament (ACL) tear (Grade $3$) vs not Complete ACL tear, \textbf{(3)} Acute ACL tears (MRI examination was performed within $6$ weeks of injury) and typified by the presence of diffuse or focal increased signal within the ligament vs not Acute ACL tear~\citep{diamond}, \textbf{(4)} Severe vs not severe neural foraminal stenosis in the cervical spine as severe
foraminal stenosis may indicate nerve impingement, which is clinically significant. Acute tear in ACL refers to the age of the tear/injury whereas the complete tear refers to the integrity of the ligament. 
\begin{table}[h]
\begin{adjustbox}{width=\columnwidth}
    \centering
    \begin{tabular}{ccccc}
       Split  & Number of examples & Min & Max & Average \\
       \toprule
       Train & $34595$ & $79$ & $6379$ &  $567$\\
       Validation & $5754$ & $44$ & $1138$ & $303$ \\
       \bottomrule
    \end{tabular}
    \end{adjustbox}
    \caption{Statistics of our meta-training and meta-validation dataset, where the min/max/average refer to min/max/average examples per label.}
    \label{tab:metatrain_stats}
\end{table}
Our testing datasets are diverse and sampled from different institutions: the knee data, lung dataset and cervical dataset is sampled from $50, 4$ and $65$ institutions respectively and our annotation process is described in Appendix~\ref{sec:annot}. Examples of these datasets can be found in figure~\ref{fig:knee_labels} (knee), figure~\ref{fig:lung_labels} (lung), and figure~\ref{fig:cervical_labels} (cervical). 
\begin{table}[h]
    \centering
    \begin{adjustbox}{width=\columnwidth}
    \begin{tabular}{@{}l l l @{}}
  Task &
  Validation Distribution &
  Test Distribution \\ 
  \toprule
  Lung Nodule  & \begin{tabular}[c]{@{}l@{}}Low Risk : $233$\\ High Risk : $30$\end{tabular} &
  \begin{tabular}[c]{@{}l@{}}Low Risk : $347$\\ High Risk : $46$\end{tabular} \\
  \begin{tabular}[c]{@{}l@{}} Knee ACL \\ Acute Tear  \end{tabular} & \begin{tabular}[c]{@{}l@{}}Normal: $258$\\ Acute Tear: $48$\end{tabular} &
  \begin{tabular}[c]{@{}l@{}}Normal : $439$\\ Acute Tear: $93$\end{tabular}  \\
  \begin{tabular}[c]{@{}l@{}} Knee ACL \\ Complete Tear \end{tabular} & \begin{tabular}[c]{@{}l@{}}Normal : $263$\\ Complete Tear : $44$\end{tabular} &
  \begin{tabular}[c]{@{}l@{}}Normal : $429$\\ Complete Tear :$103$\end{tabular} \\

 \begin{tabular}[c]{@{}l@{}} Neural Foraminal \\
  Stenosis \end{tabular} &
  \begin{tabular}[c]{@{}l@{}}Normal : $215$\\ Abnormal : $43$\end{tabular} &
  \begin{tabular}[c]{@{}l@{}}Normal : $789$\\ Abnormal : $91$\end{tabular} \\ \bottomrule
\end{tabular}
    \end{adjustbox}
    \caption{Statistics of our downstream testing datasets}
    \label{tab:test_datasets}
\end{table}

\section{Workflow}~\label{sec:workflow}
\begin{figure}[h]
\includegraphics[width=\columnwidth]{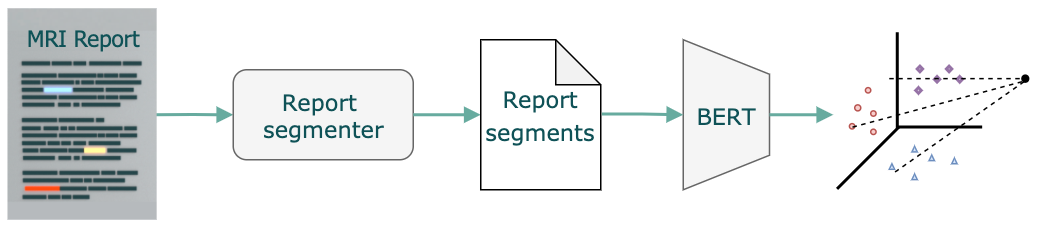}
\caption{Overview of our workflow. A report is passed through a report segmenter which splits it into sentences and extracts the relevant portion of the text for downstream classification. The relevant text is passed through our model and we use the pre-computed prototypes and class variances to assign a label to the query point.}
\label{fig:workflow}
\end{figure}
Our workflow consists of the following parts: A report is first de-identified according to HIPAA regulations and passed through a sentence parser (ex. Spacy~\citep{spacy}) that splits the report into sentences. In the shoulder dataset, each of these sentences is labeled with the appropriate structure and severity label and we filter out sentences that do not have such a label. We first train a meta-learner in an episodic fashion on this dataset and choose the best model based on meta-validation accuracy.

For our downstream tasks, we use a body-part specific custom data processor to collect sentences related to a given structure (ACL in knee, different vertebrae in the cervical spine, the entire impression section for lung reports) and concatenate them together to create a paragraph describing all the pathologies in the structure of interest. Detailed description of preprocessing for different body parts, is presented in Appendix~\ref{sec:wf}. The concatenated text from the validation sets of each task is passed to our trained meta-learner to generate the relevant class statistics (mean and variance). We then perform pathology classification on the test set by using our trained meta-learner and the saved class statistics. The downstream tasks are similar to the shoulder task in the sense that the pathology classification is performed on a sequence of sentences that all pertain to the same anatomical structure. Thus our approach needs to learn the language that describes the severity of the pathology for a specific anatomical structure. 

We would like to shed some light on the complexity of the language we encounter. Since our dataset is sourced from multiple health systems, and not all reports follow a standard structure, there is a large amount of variation in the language describing the same diagnosis. For example: a severe tear can be referred to as a rupture, or only the size of the nodule is mentioned without specifying that it is low risk (see Appendix~\ref{sec:wf} for more examples). Furthermore, most of our pipelines attempt to classify the different severities for a given pathology and the language describing severity can vary. While it might be possible to construct a rule based system to extract the diagnoses and severities we are interested in, it will be difficult to generalize as we expand to more diagnoses as well as to new health systems.

\begin{small}
\begin{table*}[t]
\begin{adjustbox}{width=.92\textwidth} 
\centering
\begin{tabular}{cccccc}
Backbone & Methods & Foraminal & \begin{tabular}[c]{@{}l@{}} Knee \\ (Acute Tear vs Not) \end{tabular} & \begin{tabular}[c]{@{}l@{}} Knee \\ (Complete tear vs Not) \end{tabular} &  Lung  \\
\toprule
& Baseline & $0.38$ & $0.44$ & $0.49$ & $0.36$ \\
& Multi-Task & $0.41$ & $0.47$ & $0.52$ & $0.39$ \\
& Vanilla ProtoNet & $0.79$ & $0.73$ & $0.60$ & $0.68$\\
& Big ProtoNet & $0.58$ & $0.59$ & $0.51$ & $0.64$\\
& Leopard & $0.84$ & $0.78$ & $0.80$ & $0.74$\\
PubMedBERT  & ProtoNet w/ Isotropic Gaussian & $0.81$ & $0.74$ & $0.76$ & $0.69$\\
& ProtoNet w/ Isotropic Gaussian + reg & $0.83$ & $0.76$ & $0.77$ & $0.73$\\
& Variance Aware ProtoNet (ours) & $0.84$ & $0.78$ & $0.79$ & $0.76$\\
& Variance Aware ProtoNet + reg (ours)& $0.86$ & $0.81$ & $0.84$ & $\mathbf{0.80}$\\
\midrule
& Baseline & $0.42$ & $0.47$ & $0.51$ & $0.41$ \\
& Multi-Task & $0.44$ & $0.49$ & $0.53$ & $0.43$ \\
& Vanilla ProtoNet & $0.78$ & $0.71$ & $0.69$ & $0.66$\\
& Big ProtoNet & $0.59$ & $0.57$ & $0.54$ & $0.67$\\
PubMedBERT w/ Adapters & ProtoNet w/ Isotropic Gaussian & $0.83$ & $0.75$ & $0.78$ & $0.72$\\
& ProtoNet w/ Isotropic Gaussian + reg & $0.89$ & $0.80$ & $0.86$ & $0.77$\\
& Variance Aware ProtoNet (ours) & $0.87$ & $0.77$ & $0.81$ & $0.74$\\
& Variance Aware ProtoNet + reg (ours)& $\mathbf{0.91}$ & $\mathbf{0.82}$& $\mathbf{0.89}$ & $0.78$\\
\bottomrule
\end{tabular}
\end{adjustbox}
\caption{Table showing F1 scores of Few Shot Models in downstream classification tasks.}
\label{tab:classificationresults}
\end{table*}
\end{small}
\section{Prototypical Networks}
Prototypical Networks or ProtoNets~\citep{protonets} use an embedding function $f_{\theta}$ to encode each input into a $M$-dimensional feature vector. A prototype is defined for every class $c \in \mathcal{L}$, as the mean of the set of embedded support data samples ($S_c$) for the given class, i.e.
\begin{equation}
v_c=\frac{1}{|S_c|}\sum_{(x_i,y_i)\in S_c}f_{\theta}(x_i).
\end{equation} 
The distribution over classes for a given test input $x$ is a softmax over the inverse of distances between the test data embedding and prototype vectors.
\begin{equation}~\label{eqn:proto}
\begin{split}
P(y=c|x) & =\text{softmax}(-d(f_{\theta}(x),v_c)) \\ 
 & =\frac{\text{exp}(-d(f_{\theta}(x),v_c))}{\sum_{c'\in L}\text{exp}(-d(f_{\theta}(x),v_{c'}))} 
\end{split}
\end{equation}
where $d$ can be any (differentiable) distance function. The loss function is negative log-likelihood: \begin{equation*}
\mathbb{L}(\theta)=-\text{log}P_{\theta}(y=c|x).
\end{equation*}
ProtoNets are simple and easy to train and deploy. The mean is used to capture the entire conditional distribution $P(y=c|x)$, thus losing a lot of information about the underlying distribution. A lot of work~\citep{ding2022fewshot,pmlr-v97-allen19b,9010263} has focused on improving ProtoNets by taking into account the above observation. We extend ProtoNets by incorporating the variance ($2$nd moment) of the distribution and use distributional distance, i.e. $2$-Wasserstein metric, directly generalizing the vanilla ProtoNets. 
\subsection{Variance Aware ProtoNets}
In this work, we model each conditional distribution as a Gaussian. Now the main question is: \textit{how do we match a query example with a distribution?} The simplest thing here is to treat the query example as a Dirac distribution. With that formulation in mind, recall: the Wasserstein-Bures metric between Gaussians $(m_i, \Sigma_i)$ is given by:
 \begin{equation*}
     d^2 = ||m_1-m_2||^2 + \text{Tr}(\Sigma_1 + \Sigma_2 - 2(\Sigma_{1}^{\frac{1}{2}}\Sigma_2\Sigma_{1}^{\frac{1}{2}})^{\frac{1}{2}})
\end{equation*}
Given $(x_i, y_i) \in S_c$, where $S_c$ is the support set of examples belonging to class $c$, we compute the mean $m_c$ and covariance matrix $\Sigma_c$; the computation of Wasserstein distance between a Gaussian and a query vector $q$ (i.e. a Dirac) boils down to 
\begin{equation}
    d^2 = ||m_c-q||^2 + \text{Tr}(\Sigma_c)
\end{equation}
The above formula shows that we can simplify our conditional distribution to be a Gaussian with a \textit{diagonal} covariance matrix. This brings down our space complexity to store this covariance matrix from O($n^2$) to O($n$). Note, this is a direct generalization of the vanilla prototypical networks as the vanilla prototypical networks can be interpreted as computing the Wasserstein distance (aka simple $L_2$ distance) between two Dirac distributions (mean of the conditional distribution and the query sample). We also propose another variant of the above called Isotropic Gaussian variant where we average over the diagonal entries of $\Sigma_c$, i.e. $\alpha = \frac{1}{n}(\Sigma_c)_{ii}$ and redefine $\Sigma_c = \alpha I$, where $I$ is the identity matrix, allowing us to just store the scalar $\alpha$, further reducing the space complexity. 
Furthermore, we regularize the negative log likelihood loss to prevent the variance term from blowing up. Our new loss function reads:
\begin{equation}
\mathcal{L}(\theta) = \mathbb{L}(\theta) +\frac{\lambda}{\text{ways}} ||\Sigma_c||_F
 \end{equation}
 where ways are the number of classes in the mini-batch and $||\cdot||_F$ is the Frobenius norm and we average the norm of the variance matrix over all the classes in a given meta-batch. The extra regularization term is designed to encourage the examples to be close to the appropriate cluster centroid. This term can also be seen as an entropic regularization term, i.e. up to a factor as the exponential of $KL(p||q)$, where $p = N(m_c, \Sigma_c)$ and $q = N(m_c, I)$. This type of entropy regularized Wasserstein distances is widely studied~\citep{pmlr-v32-cuturi14, altschuler2021averaging}. 
 
 A PyTorch style pseudocode is described in Algorithm~\ref{alg:cap}, where the {\color{teal} teal} color refers to the changes to a vanilla prototypical networks trainer. We provide detailed motivation for using Wasserstein distance instead of KL divergence in section~\ref{sec:wass_mot}. This also explains why we compute the Wasserstein distance between the query and the estimated class distribution instead of a simple likelihood. 
\section{Experiments}
All our experiments are run on $4$ V100 $16$ GB GPU using PyTorch~\citep{NEURIPS2019_9015} and HuggingFace libraries~\citep{wolf-etal-2020-transformers}. Bert-base~\citep{devlin-etal-2019-bert}, Clinical BERT~\citep{alsentzer-etal-2019-publicly} and PubMedBERT~\citep{msrbert} are used as our backbone models. Adapters~\citep{pfeiffer-etal-2020-adapterhub} are applied to each of these backbone models. While training adapter based models, the BERT weights are frozen and only the adapter weights are updated, thus requiring less resources to train. This idea is similar to~\citep{anil} in the sense that we are reusing the features from these deep pre-trained models. We compare our methods to Leopard~\citep{bansal-etal-2020-learning}, vanilla ProtoNets and big ProtoNets~\citep{ding2022fewshot}. Additional results with BERT-base and Clinical BERT backbones can be found in table~\ref{tab:metaval_add} and table~\ref{tab:classification_add}. Meta-training is done in an episodic manner using $4$-way $8$-shot and $16$-examples as support. For meta-training on the shoulder dataset, we set the variance regularizer hyperparameter to be $.1$. It is an important hyperparameter and detailed ablation study is conducted in section~\ref{sec:var_par}. Other hyperparameters and design choices are described in section~\ref{sec:hparam}.

To prevent overfitting on the test set, we choose the best model from each of these experiments based on the meta-validation accuracy and apply it to our downstream classification tasks. We note that these downstream tasks are significantly different from the few shot regime these models are trained in. Moreover for these downstream tasks, we train BERT models on each task and a multi-tasking model to provide additional baselines. 

In all our experiments, PubMedBERT consistently outperforms BERT-base and Clinical BERT by an average of $\mathbf{5}$ points and $\mathbf{3}$ points respectively. We believe the reason behind the improved performance is the domain specific vocabulary. Even though Clinical BERT is pre-trained on MIMIC-III~\citep{Johnson2016}, it still shares the same vocabulary as BERT-base. 

ProtoNet-BERT shows better performance and faster convergence rates during training and validation (Table~\ref{tab:results0}), but it is outperformed by ProtoNet-AdapterBERT which has fewer orders of magnitude of parameters to learn (Table~\ref{tab:classificationresults}). Like~\citep{wang2021gradtask} we believe that ProtoNet-BERT is more vulnerable to overfitting on the meta-training tasks than the ProtoNet-AdapterBERT. Finally, we note that even though Big ProtoNets work well on meta-validation, they fail on our downstream tasks. We hypothesize that it is due to the fact that big protonets are encouraged to have large radii which has the potential to become a bottleneck where the data distribution is highly imbalanced causing the spherical Gaussians to overlap. In fact, we have found that doing the exact opposite (i.e. constricting the norms of the covariance matrix), tends to produce better results.
\begin{table}[h!]
 \small
    \begin{center}
    \begin{adjustbox}{width=\columnwidth}
    \begin{tabular}{ccc}
  Backbone &  Methods  & Accuracy  \\
    \toprule
        &  Vanilla ProtoNet &  $89.1 \pm 1.1$  \\
       & Big ProtoNet  &  $90.8 \pm 1.2$\\
       & Leopard & $85.1 \pm 9.2$ \\
  PubMedBERT  &    ProtoNet w/ Isotropic Gaussian  & $90.2 \pm 1.4$ \\
     &   ProtoNet w/ Isotropic Gaussian + reg &  $92.1 \pm .8$ \\ 
      &  Variance Aware ProtoNet (ours) & $91.5 \pm 1.3$ \\
     &   Variance Aware ProtoNet + reg (ours) & $\mathbf{92.9 \pm .9}$ \\
     \midrule
      &  Vanilla ProtoNet &  $88.3 \pm 1.4$  \\
       & Big ProtoNets  &  $89.4 \pm 1.2$ \\
 \begin{tabular}[c]{@{}l@{}} PubMedBERT \\ w/Adapters \end{tabular} &    ProtoNet w/ Isotropic Gaussian  & $89.8 \pm 1.4$ \\
     &   ProtoNet w/ Isotropic Gaussian + reg &  $90.9 \pm .7$ \\ 
      &  Variance Aware ProtoNet (ours) & $90.5 \pm 1.3$ \\
     &   Variance Aware ProtoNet + reg (ours) & $91.2 \pm .8$ \\
\bottomrule
    \end{tabular}
    \end{adjustbox}
    \end{center}
    \caption{Results showing accuracy percentages on the meta-validation dataset. We sampled $1000$ tasks with $4$-way $8$-shot and $16$-support classification. We replicate each experiment over $10$ random seeds.}
    \label{tab:results0}
\end{table}
Finally instead of using the entire validation set to compute the class distribution, we also experiment with choosing a $k$ shots from the validation set to compute the class distribution (figure~\ref{fig:stability} in section~\ref{sec:stability}).

Our regularized Variance Aware ProtoNets with BERT-base $+$ Adapter is also validated on $13$ public datasets. For the models and datasets marked with $*$ in table~\ref{tab:pubres}, we use the results reported in~\citep{bansal-etal-2020-learning} and for those datasets, we use the code from~\citep{wang2021gradtask} to generate the results for ProtoNet with Bottleneck Adapters while the rest of the results are taken from~\citep{wang2021gradtask}. The variance regularization hyperparameter is set to $.01$ for these experiments. Our method beats Leopard by $\mathbf{5}$, $\mathbf{3}$ and $\mathbf{2}$ points on $\mathbf{4}$, $\mathbf{8}$ and $\mathbf{16}$ shots, respectively. The training details for these experiments can be found in  section~\ref{sec:benchmark}.
\begin{small}
\begin{table*}[t]
\begin{adjustbox}{width=.95\textwidth} 
\begin{tabular}{cccccccc}
Shots & Dataset & BERT* & MT-BERT* & Leopard & ProtoNet & ProtoNet+Adapter & \begin{tabular}[c]{@{}c@{}} Variance Aware ProtoNet \\ (Ours) \end{tabular}\\
\toprule
& airline & $42.76 \pm 13.50$ & $46.29 \pm 12.26$ & $54.95 \pm 11.81$& $\mathbf{65.39 \pm 12.73}$& $65.33 \pm 7.95$ & $62.67 \pm 11.18$\\
& disaster & $\mathbf{55.73 \pm 10.29}$  & $50.61 \pm 8.33$ & $51.45 \pm 4.25$ & $54.01 \pm 2.9$ & $53.48 \pm 4.76$ & $53.89 \pm 3.79$\\
& emotion & $9.20 \pm 3.22$  & $9.84 \pm 2.14$ & $11.71 \pm 2.16$ & $11.69 \pm 1.87$ & $12.52 \pm 1.32$ & $\mathbf{15.15 \pm 4.19}$ \\
& $\text{political\_audience}$ & $51.89 \pm 1.72$ & $51.53 \pm 1.80$ & $52.60 \pm 3.51$ &  $\mathbf{52.77 \pm 5.86}$  & $51.88 \pm 6.37$ & $52.5 \pm 6.45$ \\
& $\text{sentiment\_kitchen*}$ & $56.93 \pm 7.10$  & $60.53 \pm 9.25$ & $78.35 \pm 18.36$ & $62.71 \pm 9.53$  & $83.13 \pm 0.96$ & $\mathbf{84.16 \pm 1.37}$\\
& $\text{political\_bias}$ & $54.57 \pm 5.02$   & $54.66 \pm 3.74$ & $60.49 \pm 6.66$ & $58.26 \pm 10.42$ & $\mathbf{61.72 \pm 5.65}$ & $59.39 \pm 6.18$\\
4 & $\text{rating\_electronics*}$ &  $39.27 \pm 10.15$  & $41.20 \pm 10.69$  & $51.71 \pm 7.20$ & $37.40 \pm 3.72$  & $53.81 \pm 6.01$ & $\mathbf{55.49 \pm 5.42}$\\
& $\text{political\_message}$ & $15.64 \pm 2.73$  & $14.49 \pm 1.75$ & $15.69 \pm 1.57$ & $17.82 \pm 1.33$ & $\mathbf{20.98 \pm 1.69}$  &  $19.28 \pm .91$\\
& $\text{sentiment\_books*}$ & $54.81 \pm 3.75$ & $64.93 \pm 8.65$ & $82.54 \pm 1.33$  & $73.15 \pm 5.85$  & $83.88 \pm 0.55$ & $\mathbf{84.95 \pm 1.72}$\\
& $\text{rating\_books*}$ & $39.42 \pm 07.22$   & $38.97 \pm 13.27$ & $48.44 \pm 7.43$ & $54.92 \pm 6.18$ & $59.20 \pm 7.26$ & $\mathbf{66.18 \pm 7.89}$\\
& $\text{rating\_dvd*}$ & $32.22 \pm 08.72$ & $41.23 \pm 10.98$ & $49.76 \pm 9.80$ & $47.73 \pm 6.20$ & $50.20 \pm 10.26$ & $\mathbf{52.59 \pm 14.09}$\\
& $\text{rating\_kitchen}$ & $34.76 \pm 11.2$ & $36.77 \pm 10.62$ &  $50.21 \pm 9.63$ & $58.47 \pm 11.12$ & $55.99 \pm 9.85$  & $\mathbf{59.39 \pm 8.79}$\\
& $\text{scitail*}$ & $58.53 \pm 09.74$ & $63.97 \pm 14.36$ & $69.50 \pm 9.56$ &  $76.27 \pm 4.26$  & $77.84 \pm 2.61$ & $\mathbf{79.16 \pm 2.54}$\\
\rowcolor{Gray}
& Average & $41.98$ & $44.23$ & $52.11$ & $51.58$ & $56.15$ & $\mathbf{57.29}$\\
\midrule
& airline & $38.00 \pm 17.06$ & $49.81 \pm 10.86$ & $61.44 \pm 3.90$ & $69.14 \pm 4.84$ &  $\mathbf{69.37 \pm 2.46}$ & $69.31 \pm 2.43$\\
& disaster & $\mathbf{56.31 \pm 9.57}$ & $54.93 \pm 7.88$ &  $55.96 \pm 3.58$ & $54.48 \pm 3.17$ & $53.85 \pm 3.03$ & $55.19 \pm 2.77$\\
& emotion & $8.21 \pm 2.12$ & $11.21 \pm 2.11$ & $12.90 \pm 1.63$ & $13.10 \pm 2.64$ & $13.87 \pm 1.82$ & $\mathbf{15.1 \pm 3.58}$\\
& $\text{political\_audience}$  & $52.80 \pm 2.72$   & $54.34 \pm 2.88$ & $54.31 \pm 3.95$ & $\mathbf{55.17 \pm 4.28}$ & $53.08 \pm 6.08$ & $53.82 \pm 4.13$\\
& $\text{sentiment\_kitchen*}$ & $57.13 \pm 6.60$  & $69.66 \pm 8.05$  & $\mathbf{84.88 \pm 1.12}$ & $70.19 \pm 6.42$  & $83.48 \pm 0.44$ & $84.69 \pm .8$\\
& $\text{political\_bias}$ & $56.15 \pm 3.75$ & $54.79 \pm 4.19$ & $61.74 \pm 6.73$ & $63.22 \pm 1.96$ & $\mathbf{65.36 \pm 2.03}$ &  $64.09 \pm .58$\\
8 & $\text{rating\_electronics*}$ & $28.74 \pm 08.22$ & $45.41 \pm 09.49$ & $54.78 \pm 6.48$ & $43.64 \pm 7.31$ & $56.97 \pm 3.19$ & $\mathbf{60.24 \pm 2.62}$\\
& $\text{political\_message}$ & $13.38 \pm 1.74$  & $15.24 \pm 2.81$ & $18.02 \pm 2.32$ &  $20.40 \pm 1.12$ & $\mathbf{21.64 \pm 1.72}$ & $20.44 \pm 1.17$ \\
& $\text{sentiment\_books*}$ & $53.54 \pm 5.17$  & $67.38 \pm 9.78$ & $83.03 \pm 1.28$  & $75.46 \pm 6.87$ & $83.9 \pm 0.39$ & $\mathbf{84.68 \pm .85}$\\
& $\text{rating\_books*}$ & $39.55 \pm 10.01$  &  $46.77 \pm 14.12$ & $59.16 \pm 4.13$ & $ 52.13 \pm 4.79$ & $61.74 \pm 6.83$ &  $\mathbf{65.54 \pm 6.78}$\\
& $\text{rating\_dvd*}$ & $36.35 \pm 12.50$  & $45.24 \pm 9.76$ & $53.28 \pm 4.66$ &  $47.11 \pm 4.00$ & $53.25 \pm 7.47$ & $\mathbf{53.83 \pm 10.46}$\\
& $\text{rating\_kitchen}$ & $34.49 \pm 8.72$ & $47.98 \pm 9.73$ & $53.72 \pm 10.31$ & $\mathbf{57.08 \pm 11.54}$ & $56.27 \pm 10.70$ & $56.68 \pm 11.21$\\
& $\text{scitail*}$ &  $57.93 \pm 10.70$  & $68.24 \pm 10.33$ & $75.00 \pm 2.42$ & $78.27 \pm 0.98$  & $80.41 \pm 1.05$ &  $\mathbf{80.57 \pm .48}$\\
\rowcolor{Gray}
& Average & $40.97$ & $48.54$ &  $56.02$ & $53.8$ & $57.94$ & $\mathbf{58.78}$ \\
\midrule
& airline & $58.01 \pm 8.23$ & $57.25 \pm 9.90$ &  $62.15 \pm 5.56$ & $\mathbf{71.06 \pm 1.60}$ & $69.83 \pm 1.80$ & $69.9 \pm 1.06$\\
& disaster & $\mathbf{64.52 \pm 8.93}$ & $60.70 \pm 6.05$ & $61.32 \pm 2.83$ & $55.30 \pm 2.68$ & $57.38 \pm 5.25$ & $60.14 \pm 5.36$\\
& emotion & $13.43 \pm 2.51$ & $12.75 \pm 2.04$ &  $13.38 \pm 2.20$ & $12.81 \pm 1.21$ & $\mathbf{14.11 \pm 1.12}$ & $13.55 \pm 3.51$\\
& $\text{political\_audience}$ & $\mathbf{58.45 \pm 4.98}$ &  $55.14 \pm 4.57$ & $57.71 \pm 3.52$ & $56.16 \pm 2.81$ & $57.23 \pm 2.77$ & $56.36 \pm 2.29$\\
& $\text{sentiment\_kitchen*}$ &  $68.88 \pm 3.39$  & $77.37 \pm 6.74$ & $\mathbf{85.27 \pm 01.31}$ & $71.83 \pm 5.94$  & $83.72 \pm 0.30$ & $84.93 \pm .49$\\
& $\text{political\_bias}$ & $60.96 \pm 4.25$ & $60.30 \pm 3.26$ & $65.08 \pm 2.14$ & $61.98 \pm 6.89$ & $\mathbf{65.38 \pm 1.71}$ & $63.97 \pm 2.49$\\
16 & $\text{rating\_electronics*}$ & $45.48 \pm 06.13$  & $47.29 \pm 10.55$ & $58.69 \pm 2.41$ & $44.83 \pm 5.96$  & $56.62 \pm 5.62$ & $\mathbf{61.01 \pm 1.54}$\\
& $\text{political\_message}$ & $20.67 \pm 3.89$  & $19.20 \pm 2.20$ & $18.07 \pm 2.41$ & $21.36 \pm 0.86$ & $\mathbf{24.00 \pm 1.39}$ & $22.49 \pm 1.31$\\
& $\text{sentiment\_books*}$ &  $65.56 \pm 4.12$ &  $69.65 \pm 8.94$  & $83.33 \pm 0.79$ & $77.26 \pm 3.27$ &  $83.92 \pm 0.41$ & $\mathbf{84.91 \pm 0.66}$\\
& $\text{rating\_books*}$ & $43.08 \pm 11.78$  & $51.68 \pm 11.27$ & $61.02 \pm 4.19$ & $57.28 \pm 4.57$  &  $64.75 \pm 4.27$ & $\mathbf{67.34 \pm 7.52}$\\
& $\text{rating\_dvd*}$ &  $42.79 \pm 10.18$ &  $45.19 \pm 11.56 $& $53.52 \pm 4.77 $ & $48.39 \pm 3.74$ & $55.08 \pm 4.92$ & $\mathbf{56.63 \pm 6.11}$\\
& $\text{rating\_kitchen}$ & $47.94 \pm 8.28$ & $53.79 \pm 9.47$ & $57.00 \pm 8.69$ & $\mathbf{61.00 \pm 9.17}$ & $59.45 \pm 8.33$ & $58.34 \pm 11.72$ \\
& $\text{scitail*}$ & $65.66 \pm 06.82$& $75.35 \pm 04.80$ &  $77.03 \pm 1.82$  &  $78.59 \pm 0.48$  & $80.27 \pm .75$ & $\mathbf{80.89 \pm .23}$
\\
\rowcolor{Gray}
& Average & $50.42$ & $52.74$ & $57.97$ & $55.22$ & $59.36$ & $\mathbf{60.04}$\\
\bottomrule
\end{tabular}
\end{adjustbox}
    \caption{Results on some benchmark text datasets on a wide range of tasks from NLI, sentiment analysis and text classification. For the Variance Aware ProtoNet, we use BERT-base with bottleneck Adapters. For meta-training, WNLI
(m/mm), SST-$2$, QQP, RTE, MRPC, QNLI, and
SNLI datasets are used.}
    \label{tab:pubres}
\end{table*}
\end{small}
\section{Deployment}~\label{sec:deploy}
Based on the results described in table~\ref{tab:classificationresults}, we choose to deploy our regularized Variance Aware ProtoNet with Adapter-PubMedBERT. Our pipeline is deployed on AWS using a single p$3.2$x instance housed with one NVIDIA V100 GPU. The main pipeline components include \textbf{(1)} body-part specific report segmenter, \textbf{(2)} PubMedBERT backbone with adapters and \textbf{(3)} a dictionary of class prototypes and class variances, for all classes in the datasets. On inference, requests sent to the pipeline include a body part which the pipeline utilizes to load up the relevant report segmenter, class prototypes and variances. A report is then ingested by the pipeline, parsed by a sentencizer, grouped into segments according to its body part specific segmentation, and then passed to the model. Class probabilities and labels are inferred after computing the Wasserstein distance between the text embedding and the appropriate class distributions. These outputs and pipeline metadata are written out to an AWS Redshift database cluster. The entire pipeline is orchestrated in batch mode with a large enough batch size to maximize GPU capacity resulting in an average latency of $68$ms/report. 
\begin{figure}[h]
\centering
\includegraphics[width=\columnwidth]{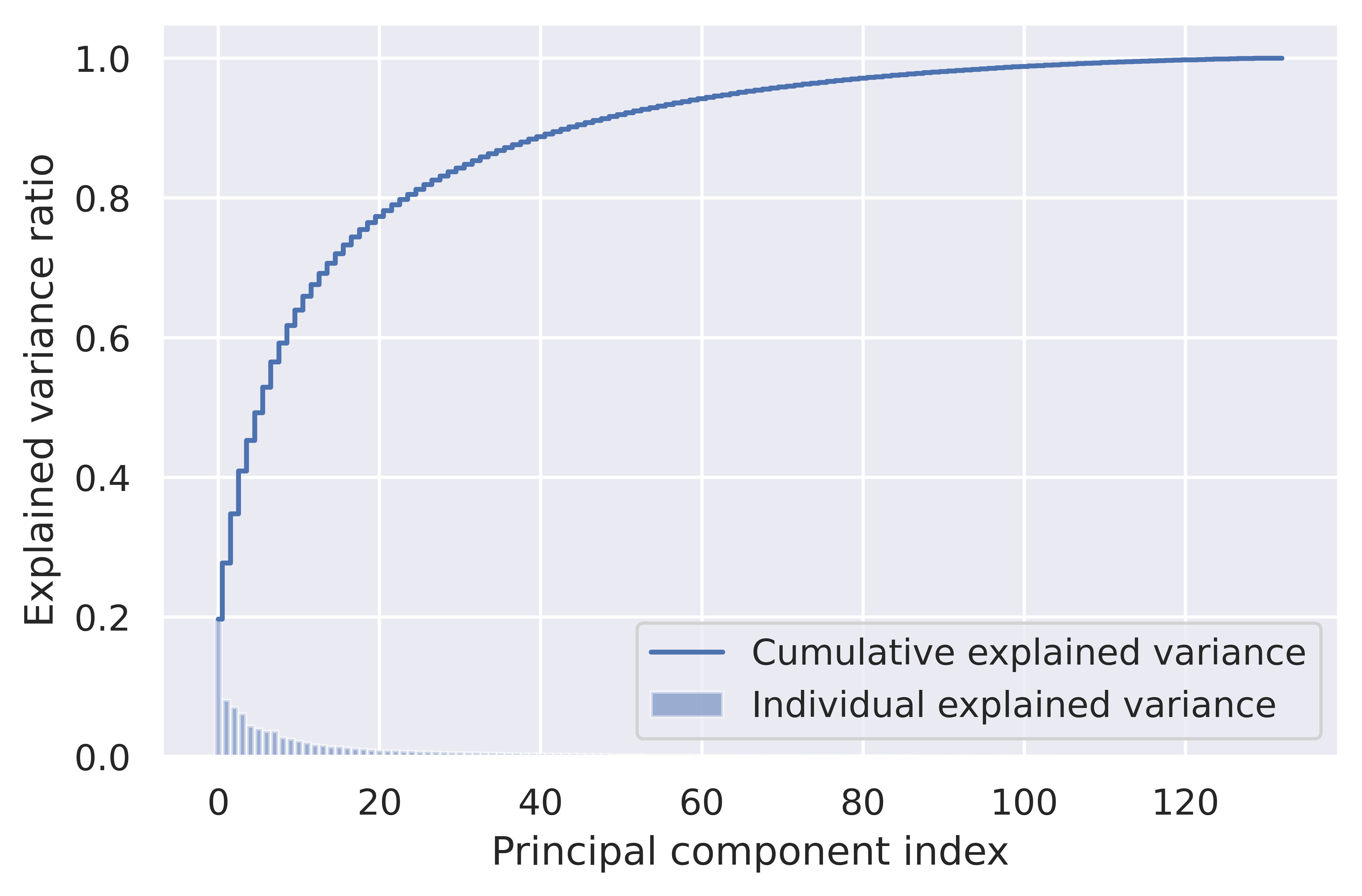}
\caption{Variance along different directions for the Lung validation set}
\label{fig:lung_pca}
\end{figure}
\subsection{Monitoring}
It is well-known that the BERT embeddings are highly anisotropic~\citep{ethayarajh-2019-contextual}. We observe the same phenomenon in our meta-learned models as well (figure~\ref{fig:lung_pca}) which we use to our advantage to monitor OOD cases. For each class in a dataset, we pick top $k$-dimensions (a hyperparameter) of maximum variance. We then take the union of these indices that we call the set of dataset indices i.e. the indices that explain the variance among all classes in the dataset. For any given query example, we compute the absolute difference ($\vec{d_j})$ between its embedding vector ($\vec{q}$) and class centroids ($\vec{v_j}$), i.e. the $i$-th coordinate $\vec{d_j}$ :  $\vec{d_j}_i = |\vec{q}_i- \vec{v_j}_i|$. We then select top $k$ dimensions of the each of these $d_j$. We propose an OOD metric called Average Variance Indices ($\text{AVI\_k}$) by the overlap between the top-k difference vector indices and the top-k dataset indices, i.e. $AVI\_k := \frac{|\cup_{j = 1}^{c} \text{top-k}(\vec{d_j})|}{\text{dataset indices}}$, where $c$ is the number of classes. For ex: in case of the lung dataset: The text \textit{"The heart is normal in size. There is no pericardial effusion. 
The pulmonary artery is enlarged."}  shows an $\text{AVI\_{10}}$ score $.79$, whereas \textit{"L1L2: There is no disc herniation in lumbar spine."} gives a score of $.31$. As part of our monitoring, we threshold reports with an $\text{AVI\_{10}} <.5$ to further investigate if the report is OOD. 
\section{Conclusion}
We extend Prototypical Networks by using Wasserstein distances instead of Euclidean distances and introduce a regularization term to encourage the class examples to be clustered close to the class prototype. By training our models on a label rich dataset (shoulder MRI reports), we show successful performance on a variety of tasks.
Since the model weights are reused for all tasks, a single model is deployed enabling us to cut inference costs. Moreover, adapters are used allowing us to tune smaller number of parameters ($\sim 10$ million) resulting in huge training cost savings. Our model is also benchmarked on $13$ public datasets and outperforms strong baselines like Leopard. Current work is underway to make our training dataset more diverse so that our models are more generalizable.

\section*{Ethical Considerations}
Due to various legal and institutional concerns arising from the sensitivity of clinical data, it is difficult for researchers to gain access to relevant data except for MIMIC~\citep{Johnson2016}. Despite
its large size (covering over $58k$ hospital admissions), it is only representative of patients from a
specific clinical domain (the intensive care unit)
and geographic location (a single hospital in the
United States). We can not expect such a sample to be representative of either the larger population of patient admissions or other geographical regions/hospital systems. We have tried to address this partially by collecting radiology data for various body parts across multiple practices in the US. However we are always mindful that our work may not generalize to new body parts/pathologies and radiology practices (see Section~\ref{sec:fails}). Even though we introduce a simple OOD metric, we realize it is far from perfect. We understand the need to minimize ethical risks of AI implementation like
threats to privacy and confidentiality, informed consent, and patient autonomy. And thus we strongly believe that stakeholders should be flexible in incorporating AI technology as
a complementary tool and not a replacement for a
physician. Thus, we develop our workflows, annotation guidelines and generate actionable insights
by working in conjunction with a varied group of
radiologists and medical professionals to minimize these above risks. Finally our pipeline as deployed is meant as a pseudo-labeling tool which we expect would cut down on expensive annotation costs but can potentially introduce some bias in our pseudo-labels.  
\bibliography{anthology,vap}

\begin{thebibliography}{41}
\expandafter\ifx\csname natexlab\endcsname\relax\def\natexlab#1{#1}\fi

\bibitem[{Allen et~al.(2019)Allen, Shelhamer, Shin, and
  Tenenbaum}]{pmlr-v97-allen19b}
Kelsey Allen, Evan Shelhamer, Hanul Shin, and Joshua Tenenbaum. 2019.
\newblock \href {https://proceedings.mlr.press/v97/allen19b.html} {Infinite
  mixture prototypes for few-shot learning}.
\newblock In \emph{Proceedings of the 36th International Conference on Machine
  Learning}, volume~97 of \emph{Proceedings of Machine Learning Research},
  pages 232--241. PMLR.

\bibitem[{Alsentzer et~al.(2019)Alsentzer, Murphy, Boag, Weng, Jindi, Naumann,
  and McDermott}]{alsentzer-etal-2019-publicly}
Emily Alsentzer, John Murphy, William Boag, Wei-Hung Weng, Di~Jindi, Tristan
  Naumann, and Matthew McDermott. 2019.
\newblock \href {https://doi.org/10.18653/v1/W19-1909} {Publicly available
  clinical {BERT} embeddings}.
\newblock In \emph{Proceedings of the 2nd Clinical Natural Language Processing
  Workshop}, pages 72--78, Minneapolis, Minnesota, USA. Association for
  Computational Linguistics.

\bibitem[{Altschuler et~al.(2021)Altschuler, Chewi, Gerber, and
  Stromme}]{altschuler2021averaging}
Jason Altschuler, Sinho Chewi, Patrik~Robert Gerber, and Austin~J Stromme.
  2021.
\newblock \href {https://openreview.net/forum?id=YV3uoawS5KK} {Averaging on the
  bures-wasserstein manifold: dimension-free convergence of gradient descent}.
\newblock In \emph{Advances in Neural Information Processing Systems}.

\bibitem[{Bansal et~al.(2020{\natexlab{a}})Bansal, Jha, and
  McCallum}]{bansal-etal-2020-learning}
Trapit Bansal, Rishikesh Jha, and Andrew McCallum. 2020{\natexlab{a}}.
\newblock \href {https://doi.org/10.18653/v1/2020.coling-main.448} {Learning to
  few-shot learn across diverse natural language classification tasks}.
\newblock In \emph{Proceedings of the 28th International Conference on
  Computational Linguistics}, pages 5108--5123, Barcelona, Spain (Online).
  International Committee on Computational Linguistics.

\bibitem[{Bansal et~al.(2020{\natexlab{b}})Bansal, Jha, Munkhdalai, and
  McCallum}]{bansal2020self}
Trapit Bansal, Rishikesh Jha, Tsendsuren Munkhdalai, and Andrew McCallum.
  2020{\natexlab{b}}.
\newblock Self-supervised meta-learning for few-shot natural language
  classification tasks.
\newblock In \emph{Proceedings of the 2020 Conference on Empirical Methods in
  Natural Language Processing (EMNLP)}, pages 522--534.

\bibitem[{Bengio et~al.(1997)Bengio, Bengio, Cloutier, and
  Gecsei}]{Bengio97onthe}
Samy Bengio, Yoshua Bengio, Jocelyn Cloutier, and Jan Gecsei. 1997.
\newblock On the optimization of a synaptic learning rule.

\bibitem[{Bowman et~al.(2015)Bowman, Angeli, Potts, and
  Manning}]{bowman-etal-2015-large}
Samuel~R. Bowman, Gabor Angeli, Christopher Potts, and Christopher~D. Manning.
  2015.
\newblock \href {https://doi.org/10.18653/v1/D15-1075} {A large annotated
  corpus for learning natural language inference}.
\newblock In \emph{Proceedings of the 2015 Conference on Empirical Methods in
  Natural Language Processing}, pages 632--642, Lisbon, Portugal. Association
  for Computational Linguistics.

\bibitem[{Brown et~al.(2020)Brown, Mann, Ryder, Subbiah, Kaplan, Dhariwal,
  Neelakantan, Shyam, Sastry, Askell, Agarwal, Herbert-Voss, Krueger, Henighan,
  Child, Ramesh, Ziegler, Wu, Winter, Hesse, Chen, Sigler, Litwin, Gray, Chess,
  Clark, Berner, McCandlish, Radford, Sutskever, and
  Amodei}]{NEURIPS2020_1457c0d6}
Tom Brown, Benjamin Mann, Nick Ryder, Melanie Subbiah, Jared~D Kaplan, Prafulla
  Dhariwal, Arvind Neelakantan, Pranav Shyam, Girish Sastry, Amanda Askell,
  Sandhini Agarwal, Ariel Herbert-Voss, Gretchen Krueger, Tom Henighan, Rewon
  Child, Aditya Ramesh, Daniel Ziegler, Jeffrey Wu, Clemens Winter, Chris
  Hesse, Mark Chen, Eric Sigler, Mateusz Litwin, Scott Gray, Benjamin Chess,
  Jack Clark, Christopher Berner, Sam McCandlish, Alec Radford, Ilya Sutskever,
  and Dario Amodei. 2020.
\newblock \href
  {https://proceedings.neurips.cc/paper/2020/file/1457c0d6bfcb4967418bfb8ac142f64a-Paper.pdf}
  {Language models are few-shot learners}.
\newblock In \emph{Advances in Neural Information Processing Systems},
  volume~33, pages 1877--1901. Curran Associates, Inc.

\bibitem[{Cuturi and Doucet(2014)}]{pmlr-v32-cuturi14}
Marco Cuturi and Arnaud Doucet. 2014.
\newblock \href {https://proceedings.mlr.press/v32/cuturi14.html} {Fast
  computation of wasserstein barycenters}.
\newblock In \emph{Proceedings of the 31st International Conference on Machine
  Learning}, volume~32 of \emph{Proceedings of Machine Learning Research},
  pages 685--693, Bejing, China. PMLR.

\bibitem[{Devlin et~al.(2019)Devlin, Chang, Lee, and
  Toutanova}]{devlin-etal-2019-bert}
Jacob Devlin, Ming-Wei Chang, Kenton Lee, and Kristina Toutanova. 2019.
\newblock \href {https://doi.org/10.18653/v1/N19-1423} {{BERT}: Pre-training of
  deep bidirectional transformers for language understanding}.
\newblock In \emph{Proceedings of the 2019 Conference of the North {A}merican
  Chapter of the Association for Computational Linguistics: Human Language
  Technologies, Volume 1 (Long and Short Papers)}, pages 4171--4186,
  Minneapolis, Minnesota. Association for Computational Linguistics.

\bibitem[{Dimond et~al.(1998)Dimond, Fadale, Hulstyn, Tung, and
  Greisberg}]{diamond}
P~Dimond, Paul Fadale, Michael Hulstyn, Glenn Tung, and J~Greisberg. 1998.
\newblock A comparison of mri findings in patients with acute and chronic acl
  tears.
\newblock \emph{The American journal of knee surgery}, 11:153--9.

\bibitem[{Ding et~al.(2022)Ding, Chen, Wang, Zheng, Liu, and
  Xie}]{ding2022fewshot}
Ning Ding, Yulin Chen, Xiaobin Wang, Hai-Tao Zheng, Zhiyuan Liu, and Pengjun
  Xie. 2022.
\newblock \href {https://openreview.net/forum?id=mL07kYPn3E} {Few-shot learning
  with big prototypes}.

\bibitem[{Ethayarajh(2019)}]{ethayarajh-2019-contextual}
Kawin Ethayarajh. 2019.
\newblock \href {https://doi.org/10.18653/v1/D19-1006} {How contextual are
  contextualized word representations? {C}omparing the geometry of {BERT},
  {ELM}o, and {GPT}-2 embeddings}.
\newblock In \emph{Proceedings of the 2019 Conference on Empirical Methods in
  Natural Language Processing and the 9th International Joint Conference on
  Natural Language Processing (EMNLP-IJCNLP)}, pages 55--65, Hong Kong, China.
  Association for Computational Linguistics.

\bibitem[{Finn et~al.(2017)Finn, Abbeel, and Levine}]{pmlr-v70-finn17a}
Chelsea Finn, Pieter Abbeel, and Sergey Levine. 2017.
\newblock \href {https://proceedings.mlr.press/v70/finn17a.html}
  {Model-agnostic meta-learning for fast adaptation of deep networks}.
\newblock In \emph{Proceedings of the 34th International Conference on Machine
  Learning}, volume~70 of \emph{Proceedings of Machine Learning Research},
  pages 1126--1135. PMLR.

\bibitem[{Ge et~al.(2022)Ge, Guo, Yang, Al-Garadi, and Sarker}]{metanlp}
Yao Ge, Yuting Guo, Yuan-Chi Yang, Mohammed~Ali Al-Garadi, and Abeed Sarker.
  2022.
\newblock \href {https://doi.org/10.48550/ARXIV.2204.14081} {Few-shot learning
  for medical text: A systematic review}.

\bibitem[{Geng et~al.(2019)Geng, Li, Li, Zhu, Jian, and Sun}]{fewshottext}
Ruiying Geng, Binhua Li, Yongbin Li, Xiaodan Zhu, Ping Jian, and Jian Sun.
  2019.
\newblock \href {https://doi.org/10.48550/ARXIV.1902.10482} {Induction networks
  for few-shot text classification}.

\bibitem[{Gu et~al.(2021)Gu, Tinn, Cheng, Lucas, Usuyama, Liu, Naumann, Gao,
  and Poon}]{msrbert}
Yu~Gu, Robert Tinn, Hao Cheng, Michael Lucas, Naoto Usuyama, Xiaodong Liu,
  Tristan Naumann, Jianfeng Gao, and Hoifung Poon. 2021.
\newblock \href {https://doi.org/10.1145/3458754} {Domain-specific language
  model pretraining for biomedical natural language processing}.
\newblock \emph{ACM Trans. Comput. Healthcare}, 3(1).

\bibitem[{He et~al.(2022)He, Zhou, Ma, Berg-Kirkpatrick, and Neubig}]{mam}
Junxian He, Chunting Zhou, Xuezhe Ma, Taylor Berg-Kirkpatrick, and Graham
  Neubig. 2022.
\newblock \href {https://openreview.net/forum?id=0RDcd5Axok} {Towards a unified
  view of parameter-efficient transfer learning}.
\newblock In \emph{International Conference on Learning Representations}.

\bibitem[{Honnibal et~al.(2020)Honnibal, Montani, Van~Landeghem, and
  Boyd}]{spacy}
Matthew Honnibal, Ines Montani, Sofie Van~Landeghem, and Adriane Boyd. 2020.
\newblock \href {https://doi.org/10.5281/zenodo.1212303} {{spaCy:
  Industrial-strength Natural Language Processing in Python}}.

\bibitem[{Hospedales et~al.(2020)Hospedales, Antoniou, Micaelli, and
  Storkey}]{metalearning_survey}
Timothy Hospedales, Antreas Antoniou, Paul Micaelli, and Amos Storkey. 2020.
\newblock \href {https://doi.org/10.48550/ARXIV.2004.05439} {Meta-learning in
  neural networks: A survey}.

\bibitem[{Johnson et~al.(2016)Johnson, Pollard, Shen, Lehman, Feng, Ghassemi,
  Moody, Szolovits, Anthony~Celi, and Mark}]{Johnson2016}
Alistair~E.W. Johnson, Tom~J. Pollard, Lu~Shen, Li-wei~H. Lehman, Mengling
  Feng, Mohammad Ghassemi, Benjamin Moody, Peter Szolovits, Leo Anthony~Celi,
  and Roger~G. Mark. 2016.
\newblock \href {https://doi.org/10.1038/sdata.2016.35} {Mimic-iii, a freely
  accessible critical care database}.
\newblock \emph{Scientific Data}, 3(1):160035.

\bibitem[{Karimi~Mahabadi et~al.(2021)Karimi~Mahabadi, Ruder, Dehghani, and
  Henderson}]{compacter}
Rabeeh Karimi~Mahabadi, Sebastian Ruder, Mostafa Dehghani, and James Henderson.
  2021.
\newblock Parameter-efficient multi-task fine-tuning for transformers via
  shared hypernetworks.
\newblock In \emph{Annual Meeting of the Association for Computational
  Linguistics}.

\bibitem[{Lake et~al.(2015{\natexlab{a}})Lake, Salakhutdinov, and
  Tenenbaum}]{learningtolearn}
Brenden~M. Lake, Ruslan Salakhutdinov, and Joshua~B. Tenenbaum.
  2015{\natexlab{a}}.
\newblock \href {https://doi.org/10.1126/science.aab3050} {Human-level concept
  learning through probabilistic program induction}.
\newblock \emph{Science}, 350(6266):1332--1338.

\bibitem[{Lake et~al.(2015{\natexlab{b}})Lake, Salakhutdinov, and
  Tenenbaum}]{omniglot}
Brenden~M. Lake, Ruslan Salakhutdinov, and Joshua~B. Tenenbaum.
  2015{\natexlab{b}}.
\newblock \href {https://doi.org/10.1126/science.aab3050} {Human-level concept
  learning through probabilistic program induction}.
\newblock \emph{Science}, 350(6266):1332--1338.

\bibitem[{Nair et~al.(2018)Nair, Devaraj, Callister, and Baldwin}]{Nair806}
Arjun Nair, Anand Devaraj, Matthew E~J Callister, and David~R Baldwin. 2018.
\newblock \href {https://doi.org/10.1136/thoraxjnl-2018-211764} {The fleischner
  society 2017 and british thoracic society 2015 guidelines for managing
  pulmonary nodules: keep calm and carry on}.
\newblock \emph{Thorax}, 73(9):806--812.

\bibitem[{Nichol et~al.(2018)Nichol, Achiam, and Schulman}]{reptile}
Alex Nichol, Joshua Achiam, and John Schulman. 2018.
\newblock \href {https://doi.org/10.48550/ARXIV.1803.02999} {On first-order
  meta-learning algorithms}.

\bibitem[{Paszke et~al.(2019)Paszke, Gross, Massa, Lerer, Bradbury, Chanan,
  Killeen, Lin, Gimelshein, Antiga, Desmaison, Kopf, Yang, DeVito, Raison,
  Tejani, Chilamkurthy, Steiner, Fang, Bai, and Chintala}]{NEURIPS2019_9015}
Adam Paszke, Sam Gross, Francisco Massa, Adam Lerer, James Bradbury, Gregory
  Chanan, Trevor Killeen, Zeming Lin, Natalia Gimelshein, Luca Antiga, Alban
  Desmaison, Andreas Kopf, Edward Yang, Zachary DeVito, Martin Raison, Alykhan
  Tejani, Sasank Chilamkurthy, Benoit Steiner, Lu~Fang, Junjie Bai, and Soumith
  Chintala. 2019.
\newblock \href
  {http://papers.neurips.cc/paper/9015-pytorch-an-imperative-style-high-performance-deep-learning-library.pdf}
  {Pytorch: An imperative style, high-performance deep learning library}.
\newblock In H.~Wallach, H.~Larochelle, A.~Beygelzimer, F.~d\textquotesingle
  Alch\'{e}-Buc, E.~Fox, and R.~Garnett, editors, \emph{Advances in Neural
  Information Processing Systems 32}, pages 8024--8035. Curran Associates, Inc.

\bibitem[{Pfeiffer et~al.(2020)Pfeiffer, R{\"u}ckl{\'e}, Poth, Kamath,
  Vuli{\'c}, Ruder, Cho, and Gurevych}]{pfeiffer-etal-2020-adapterhub}
Jonas Pfeiffer, Andreas R{\"u}ckl{\'e}, Clifton Poth, Aishwarya Kamath, Ivan
  Vuli{\'c}, Sebastian Ruder, Kyunghyun Cho, and Iryna Gurevych. 2020.
\newblock \href {https://doi.org/10.18653/v1/2020.emnlp-demos.7}
  {{A}dapter{H}ub: A framework for adapting transformers}.
\newblock In \emph{Proceedings of the 2020 Conference on Empirical Methods in
  Natural Language Processing: System Demonstrations}, pages 46--54, Online.
  Association for Computational Linguistics.

\bibitem[{Raghu et~al.(2019)Raghu, Raghu, Bengio, and Vinyals}]{anil}
Aniruddh Raghu, Maithra Raghu, Samy Bengio, and Oriol Vinyals. 2019.
\newblock \href {https://doi.org/10.48550/ARXIV.1909.09157} {Rapid learning or
  feature reuse? towards understanding the effectiveness of maml}.

\bibitem[{Roelofs et~al.(2019)Roelofs, Shankar, Recht, Fridovich-Keil, Hardt,
  Miller, and Schmidt}]{metaoverfitting}
Rebecca Roelofs, Vaishaal Shankar, Benjamin Recht, Sara Fridovich-Keil, Moritz
  Hardt, John Miller, and Ludwig Schmidt. 2019.
\newblock \href
  {https://proceedings.neurips.cc/paper/2019/file/ee39e503b6bedf0c98c388b7e8589aca-Paper.pdf}
  {A meta-analysis of overfitting in machine learning}.
\newblock In \emph{Advances in Neural Information Processing Systems},
  volume~32. Curran Associates, Inc.

\bibitem[{Schmidhuber(1987)}]{schmidhuber:1987:srl}
Jurgen Schmidhuber. 1987.
\newblock \href {http://www.idsia.ch/~juergen/diploma.html} {Evolutionary
  principles in self-referential learning. on learning now to learn: The
  meta-meta-meta...-hook}.
\newblock Diploma thesis, Technische Universitat Munchen, Germany, 14 May.

\bibitem[{Sehanobish et~al.(2022)Sehanobish, Sandora, Abraham, Pawar, Torres,
  Das, Becker, Herzog, Odry, and Vianu}]{naacl_spine}
Arijit Sehanobish, McCullen Sandora, Nabila Abraham, Jayashri Pawar, Danielle
  Torres, Anasuya Das, Murray Becker, Richard Herzog, Benjamin Odry, and Ron
  Vianu. 2022.
\newblock \href {https://aclanthology.org/2022.naacl-industry.16} {Explaining
  the effectiveness of multi-task learning for efficient knowledge extraction
  from spine {MRI} reports}.
\newblock In \emph{Proceedings of the 2022 Conference of the North American
  Chapter of the Association for Computational Linguistics: Human Language
  Technologies: Industry Track}, pages 130--140, Hybrid: Seattle, Washington +
  Online. Association for Computational Linguistics.

\bibitem[{Singh et~al.(2021)Singh, Bharti, Purohit, Kumar, Singh, and
  Singh}]{SINGH2021108111}
Rishav Singh, Vandana Bharti, Vishal Purohit, Abhinav Kumar, Amit~Kumar Singh,
  and Sanjay~Kumar Singh. 2021.
\newblock \href {https://doi.org/https://doi.org/10.1016/j.patcog.2021.108111}
  {Metamed: Few-shot medical image classification using gradient-based
  meta-learning}.
\newblock \emph{Pattern Recognition}, 120:108111.

\bibitem[{Snell et~al.(2017)Snell, Swersky, and Zemel}]{protonets}
Jake Snell, Kevin Swersky, and Richard Zemel. 2017.
\newblock Prototypical networks for few-shot learning.
\newblock In \emph{Proceedings of the 31st International Conference on Neural
  Information Processing Systems}, NIPS'17, page 4080–4090, Red Hook, NY,
  USA. Curran Associates Inc.

\bibitem[{Swartz et~al.(2005)Swartz, Floyd, and Cendoma}]{Swartz2005}
Erik~E. Swartz, R.~T. Floyd, and Mike Cendoma. 2005.
\newblock \href {https://pubmed.ncbi.nlm.nih.gov/16284634} {{Cervical Spine
  Functional Anatomy and the Biomechanics of Injury due to Compressive
  Loading}}.
\newblock \emph{Journal of athletic training}, 40(3):155--161.

\bibitem[{Vinyals et~al.(2016)Vinyals, Blundell, Lillicrap, kavukcuoglu, and
  Wierstra}]{matchingnet}
Oriol Vinyals, Charles Blundell, Timothy Lillicrap, koray kavukcuoglu, and Daan
  Wierstra. 2016.
\newblock \href
  {https://proceedings.neurips.cc/paper/2016/file/90e1357833654983612fb05e3ec9148c-Paper.pdf}
  {Matching networks for one shot learning}.
\newblock In \emph{Advances in Neural Information Processing Systems},
  volume~29. Curran Associates, Inc.

\bibitem[{Wah et~al.(2011)Wah, Branson, Welinder, Perona, and
  Belongie}]{WahCUB_200_2011}
C.~Wah, S.~Branson, P.~Welinder, P.~Perona, and S.~Belongie. 2011.
\newblock Caltech-ucsd birds-200-2011.
\newblock Technical Report CNS-TR-2011-001, California Institute of Technology.

\bibitem[{Wang et~al.(2021)Wang, Wang, Rudzicz, and Brudno}]{wang2021gradtask}
Jixuan Wang, Kuan-Chieh Wang, Frank Rudzicz, and Michael Brudno. 2021.
\newblock \href {https://openreview.net/forum?id=jScy7BjbZeQ} {Grad2task:
  Improved few-shot text classification using gradients for task
  representation}.
\newblock In \emph{Advances in Neural Information Processing Systems}.

\bibitem[{Wolf et~al.(2020)Wolf, Debut, Sanh, Chaumond, Delangue, Moi, Cistac,
  Rault, Louf, Funtowicz, Davison, Shleifer, von Platen, Ma, Jernite, Plu, Xu,
  Scao, Gugger, Drame, Lhoest, and Rush}]{wolf-etal-2020-transformers}
Thomas Wolf, Lysandre Debut, Victor Sanh, Julien Chaumond, Clement Delangue,
  Anthony Moi, Pierric Cistac, Tim Rault, Rémi Louf, Morgan Funtowicz, Joe
  Davison, Sam Shleifer, Patrick von Platen, Clara Ma, Yacine Jernite, Julien
  Plu, Canwen Xu, Teven~Le Scao, Sylvain Gugger, Mariama Drame, Quentin Lhoest,
  and Alexander~M. Rush. 2020.
\newblock \href {https://www.aclweb.org/anthology/2020.emnlp-demos.6}
  {Transformers: State-of-the-art natural language processing}.
\newblock In \emph{Proceedings of the 2020 Conference on Empirical Methods in
  Natural Language Processing: System Demonstrations}, pages 38--45, Online.
  Association for Computational Linguistics.

\bibitem[{Yogatama et~al.(2019)Yogatama, d'Autume, Connor, Kocisky,
  Chrzanowski, Kong, Lazaridou, Ling, Yu, Dyer, and Blunsom}]{finetuning}
Dani Yogatama, Cyprien de~Masson d'Autume, Jerome Connor, Tomas Kocisky, Mike
  Chrzanowski, Lingpeng Kong, Angeliki Lazaridou, Wang Ling, Lei Yu, Chris
  Dyer, and Phil Blunsom. 2019.
\newblock \href {https://doi.org/10.48550/ARXIV.1901.11373} {Learning and
  evaluating general linguistic intelligence}.

\bibitem[{Zhang et~al.(2019)Zhang, Zhao, Ni, Xu, and Yang}]{9010263}
Jian Zhang, Chenglong Zhao, Bingbing Ni, Minghao Xu, and Xiaokang Yang. 2019.
\newblock \href {https://doi.org/10.1109/ICCV.2019.00177} {Variational few-shot
  learning}.
\newblock In \emph{2019 IEEE/CVF International Conference on Computer Vision
  (ICCV)}, pages 1685--1694.

\end{thebibliography}
\bibliographystyle{acl_natbib}

\appendix

\section{Annotation}~\label{sec:annot}
First we collect data from various sources and a part of the data are annotated by our team of in-house expert annotators with deep clinical expertise, which we use as test and development sets for our model training. We then use this annotated data to train a larger pool of other annotators who are generally medical students. They are provided clear guidelines on the task and performance is measured periodically on a benchmark set and feedback is provided. As of the writing of the manuscript, the validation and the test sets as described in section~\ref{sec:datasets} are being used to train the annotators. After the completion of their training, the annotators will annotate the remaining unlabeled data that will be used as a training data for our models. The entire process is slow but is designed to generate high quality annotated data. We believe that our few shot models can be used as a source of pseudo-labels and will greatly simplify and quicken our annotation process. 

\section{Shoulder Dataset}~\label{sec:metatrain}
\begin{figure}[h]
\includegraphics[page=1,width=\columnwidth]{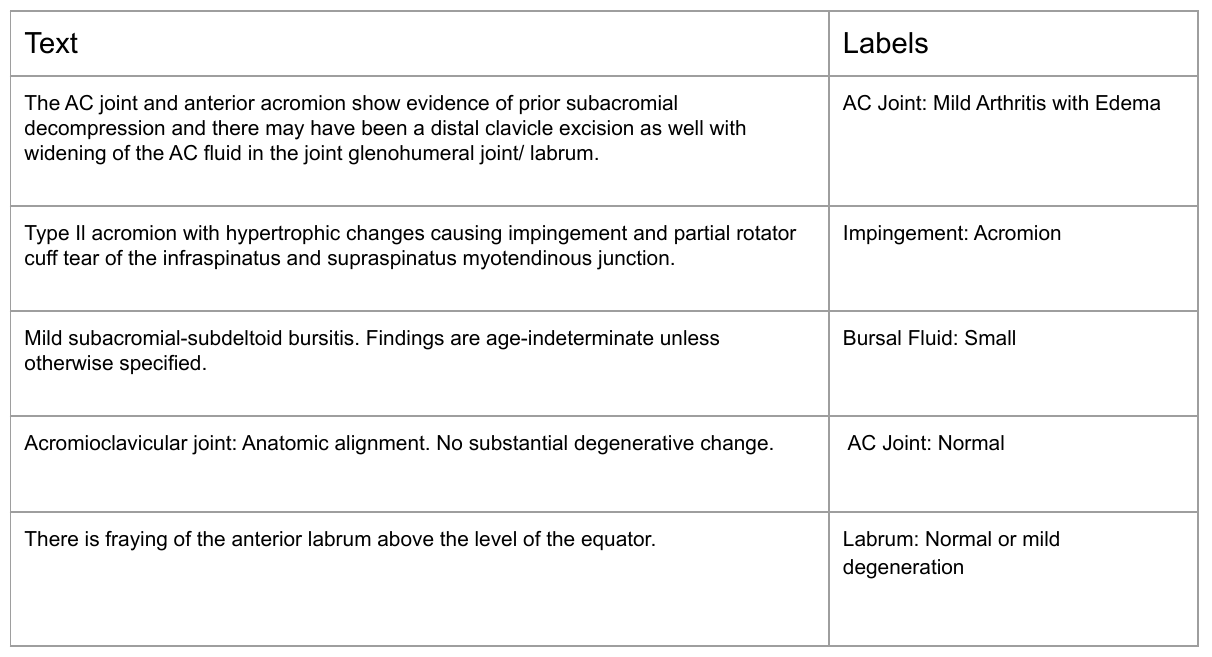}
\caption{Figure showing an example of our shoulder dataset which is used for meta-training. Note that the labels attached to the text have information about the location and severity of a given pathology.}
\label{fig:shoulder_example}
\end{figure}

In this section we will briefly describe our label rich shoulder dataset that is used as meta-training and meta-validation sets. There are $80$ labels for the shoulder dataset. They range from Clinical history, metadata, Impressions, Finding to various granular pathologies at different structures in the shoulder like AC joint, Rotator Cuff, Muscles, Bursal Fluid, Supraspinatus, Infraspinatus, Subscapularis, Labrum, Glenohumeral Joint, Humeral Head, Acromial Morphology, Impingement: AC Joint. The labels are split such that all pathologies in a given structure appear at either training or validation but not both. We believe that such a split would help a model to learn the key words that may describe the granularity of a pathology in a given structure of interest. The dataset level statistics can be found in figure~\ref{fig:histplot} and table~\ref{tab:metatrain_stats}. An example of the shoulder data is shown in figure~\ref{fig:shoulder_example}.

\section{Detailed Workflow}~\label{sec:wf}
We now present a detailed description of various body part specific workflows. All reports, irrespective of body part, are first de-identified according to HIPAA regulations. We then pass the report through a sentence parser to parse the report in sentences. 
\subsection{Lung Dataset}
\begin{figure}[h]
\includegraphics[width=\columnwidth]{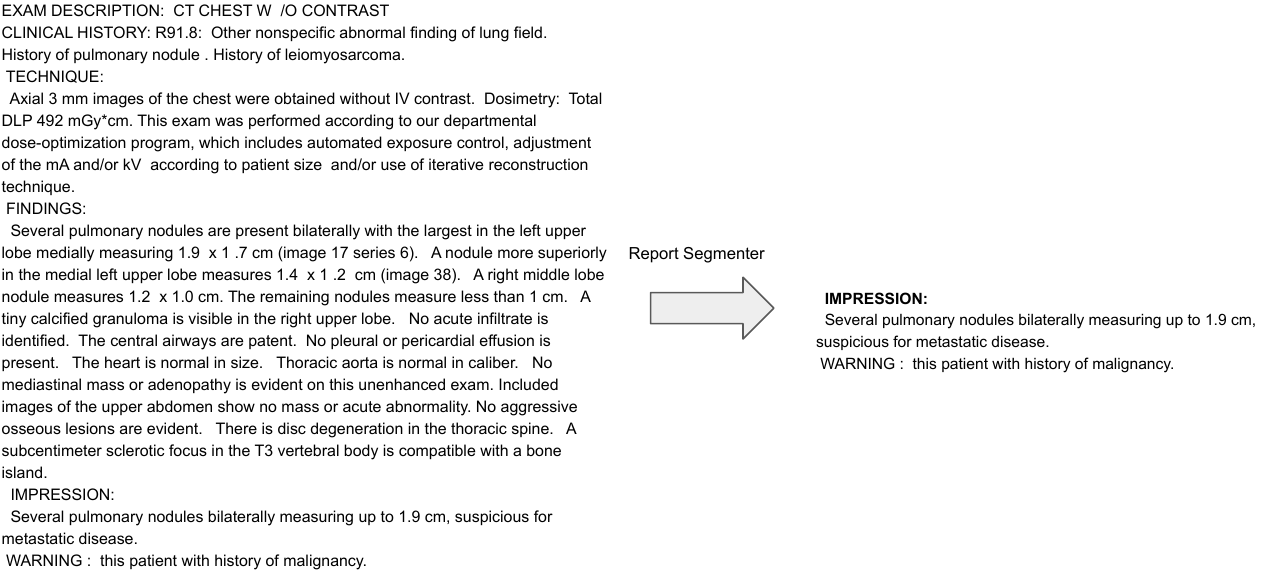}
\caption{Figure showing the preprocessing of the lung dataset. Our report segmenter extracts the relevant paragraph which is used for downstream classification.}
\label{fig: lung_wf}
\end{figure}

\begin{figure}[h]
    \centering
    \includegraphics[page=2,width=\columnwidth]{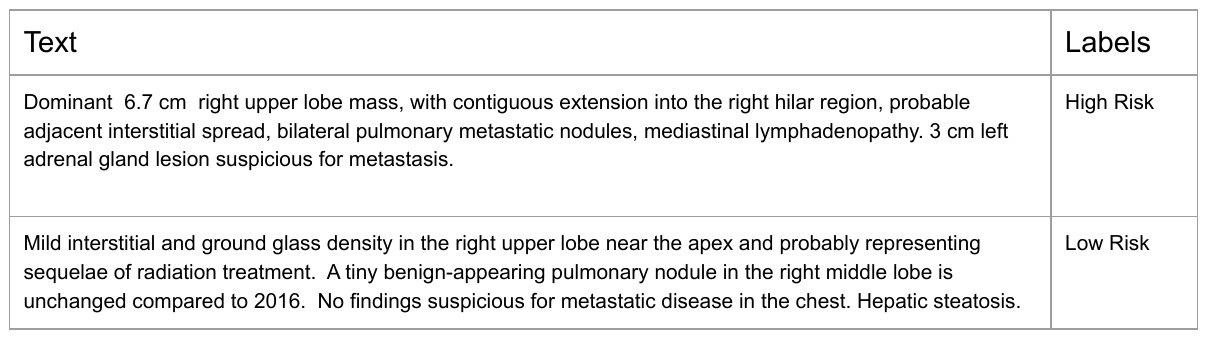}
    \caption{Figure showing the labels in the Lung dataset}
    \label{fig:lung_labels}
\end{figure}

For the lung dataset, we use a report segmenter which is a rule-based regex to extract the ``Impression" section from the entire report. This section can be thought as the summary of the report and contains all the critical information like number of lung nodules and their sizes and potential for malignancy. This section text is used for final classification task as shown in figure~\ref{fig: lung_wf}. Figure~\ref{fig:lung_labels} shows examples of the labels in the dataset. 

\subsection{Cervical Dataset}
\begin{figure}[h]
\includegraphics[width=\columnwidth]{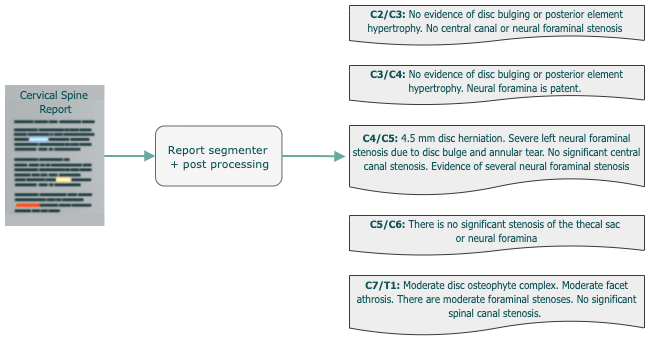}
\caption{Figure showing the preprocessing of the cervical dataset. Our report segmenter extracts all the motion segments mentioned in the report and groups all sentences belonging to the relevant motion segment. The paragraph belonging to a given motion segment is used for downstream classification.}
\label{fig: cervical_wf}
\end{figure}

\begin{figure}[h]
    \centering
    \includegraphics[page=4,width=\columnwidth]{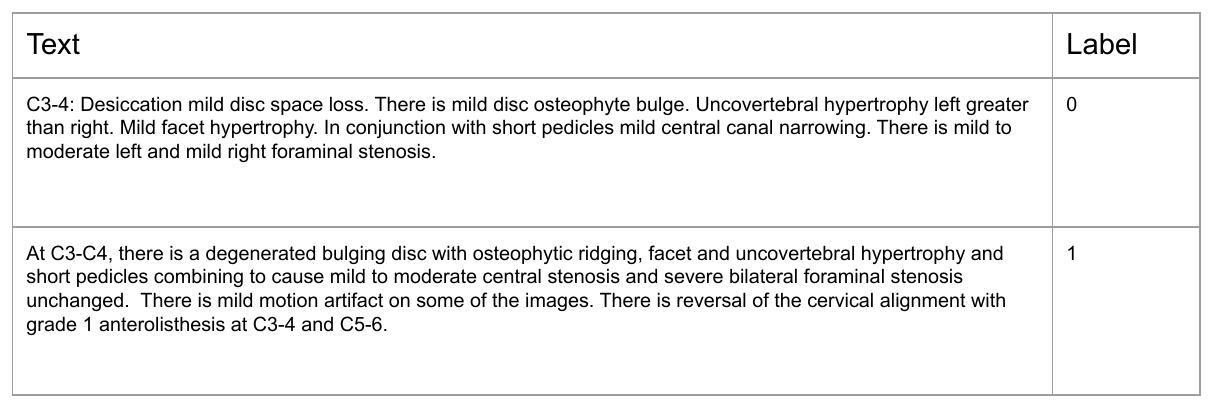}
    \caption{Figure showing the labels in the Cervical dataset. $0$ means absence of severe neural foraminal stenosis and $1$ indicates presence of severe neural foraminal stenosis.}
    \label{fig:cervical_labels}
\end{figure}

Our task in the cervical dataset is to predict the severity of a neural foraminal stenosis for each motion segment - the smallest physiological motion unit of the spinal cord~\citep{Swartz2005}. Breaking information down at the motion segment level in this way enables pathological findings to be correlated with clinical exam findings, and can inform future treatment interventions. A BERT based NER model is used to identify the motion segment(s) referenced in each sentence, and all the sentences containing a particular motion segment are concatenated together. We also use additional rule-based logic to assign motion segments to relevant sentences that may not mention a motion segment in it. We then predict the disease severity using this concatenated text at \textit{each} motion segment. This data pre-processing mostly follows the ideas and the steps outlined in~\citep{naacl_spine}. Figure~\ref{fig: cervical_wf} shows our preprocessing steps and figure~\ref{fig:cervical_labels} shows examples in the datasets.

\subsection{Knee Dataset}
\begin{figure}[h]
\includegraphics[width=\columnwidth]{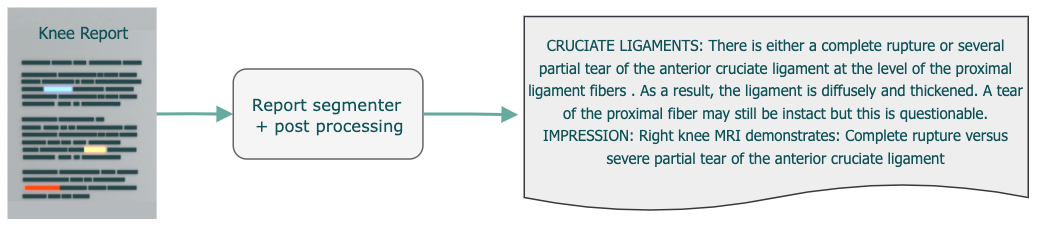}
\caption{Figure showing the preprocessing of the knee dataset. Our report segmenter selects all the relevant sentences pertaining to the structure of interest, i.e. ACL. We then predict various pathology severities using this paragraph of text.}
\label{fig: knee_wf}
\end{figure}

\begin{figure}[h]
    \centering
    \includegraphics[page=3,width=\columnwidth]{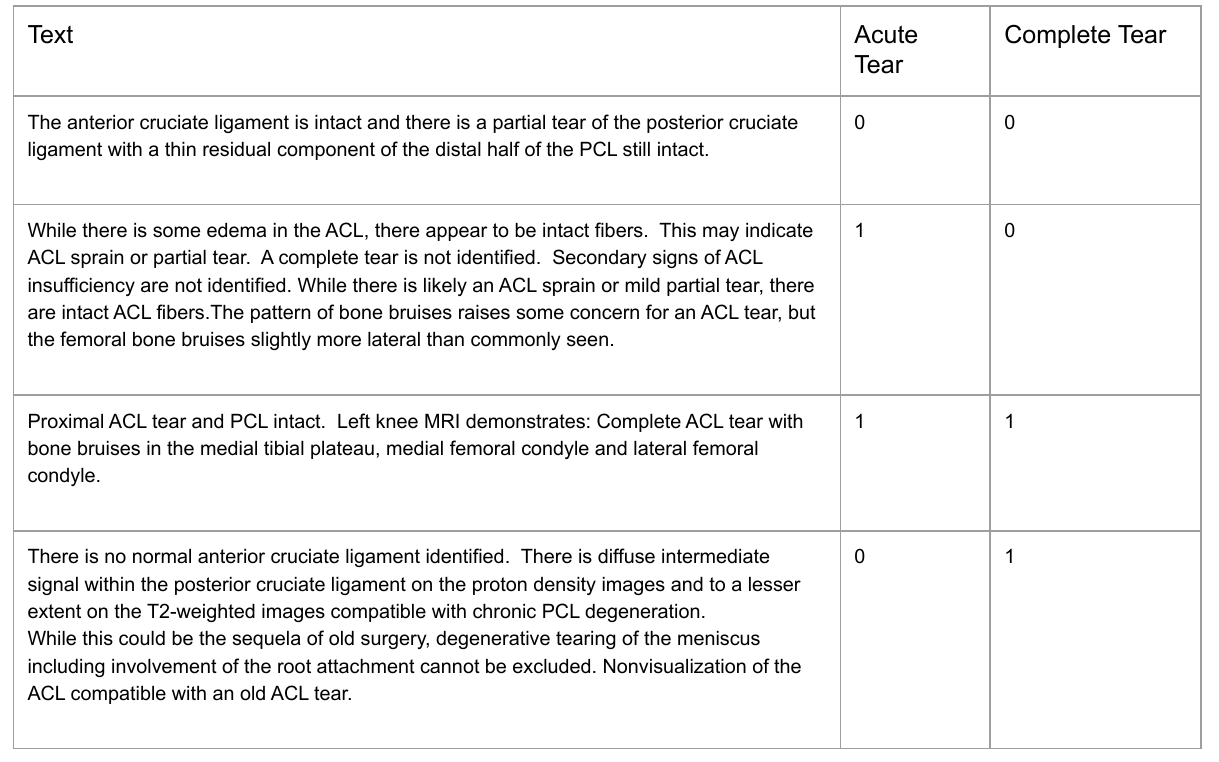}
    \caption{Figure showing the labels in the Knee dataset. $0$ means absence of a given pathology and $1$ indicates presence of such.}
    \label{fig:knee_labels}
\end{figure}

The data processing steps for the knee dataset is  similar to the cervical dataset. A BERT based NER model is used to tag sentences that mention the structure of importance, i.e. the anterior cruciate ligament (ACL). We group all the sentences together that mention ACL and we use these grouped sentences to predict our pathology severity as shown in the workflow (figure~\ref{fig: knee_wf}). An example of the labels in the knee dataset can be found in figure~\ref{fig:knee_labels}. 

\section{Additional Experiments}~\label{sec:add_expt}
We also experiment with BERT-base and Clinical BERT as additional backbones. We add adapters to these backbones as well. Finally, we choose the best model based on meta-validation accuracy and use it for our downstream tasks. In all our experiments, PubMedBERT-based backbones outperform the BERT-base and the Clinical BERT backbones.
\begin{table}[h!]
 \small
    \begin{center}
    \begin{adjustbox}{width=.9\columnwidth}
    \begin{tabular}{ccc}
  Backbone &  Methods  & Accuracy  \\
    \toprule
    &  Vanilla ProtoNet &  $86.3 \pm 1.2$ \\
    & Big ProtoNet  &  $87.8 \pm .9$ \\
    & Leopard & $81.4 \pm 9.7$ \\
  BERT-base  &    ProtoNet w/ Isotropic Gaussian  & $88.7 \pm 1.4$ \\
     &   ProtoNet w/ Isotropic Gaussian + reg &  $89.5 \pm .8$ \\ 
      &  Variance Aware ProtoNet (ours) & $88.9 \pm 1.5$ \\
     &   Variance Aware ProtoNet + reg (ours) & $90.1 \pm .9$ \\
     \midrule
             &  Vanilla ProtoNet &  $85.6 \pm 1.3$ \\
      & Big ProtoNet  &  $87.1 \pm 1.1$ \\
 \begin{tabular}[c]{@{}l@{}} BERT-base \\ w/Adapters \end{tabular}  &    ProtoNet w/ Isotropic Gaussian  & $87.8 \pm .8$ \\
     &   ProtoNet w/ Isotropic Gaussian + reg &  $88.6 \pm .7$ \\ 
      &  Variance Aware ProtoNet (ours) & $88.1 \pm 1.2$ \\
     &   Variance Aware ProtoNet + reg (ours) & $89.7 \pm .8$ \\
     \midrule
             &  Vanilla ProtoNet & $87.4 \pm  1.3$  \\
      & Big ProtoNet  &  $88.5 \pm 1.1$ \\
      & Leopard & $82.2 \pm 9.8$ \\
  Clinical BERT  &    ProtoNet w/ Isotropic Gaussian  & $89.6 \pm 1.2 $ \\
     &   ProtoNet w/ Isotropic Gaussian + reg &  $90.1 \pm .8 $ \\ 
      &  Variance Aware ProtoNet (ours) & $89.9 \pm 1.1$ \\
     &   Variance Aware ProtoNet + reg (ours) & $90.9 \pm .8$ \\
     \midrule
             &  Vanilla ProtoNet &  $86.8 \pm .9$  \\
      & Big ProtoNet  &  $87.9 \pm 1.1$ \\
 \begin{tabular}[c]{@{}l@{}} Clinical BERT \\ w/Adapters \end{tabular}  &    ProtoNet w/ Isotropic Gaussian  & $88.4 \pm 1.3 $ \\
     &   ProtoNet w/ Isotropic Gaussian + reg &  $89.1 \pm .9$ \\ 
      &  Variance Aware ProtoNet (ours) & $88.7 \pm 1.1$ \\
     &   Variance Aware ProtoNet + reg (ours) & $89.5 \pm .9$ \\
\bottomrule
\end{tabular}
\end{adjustbox}
\caption{Results showing accuracy percentages on the meta-validation dataset. We sample $1000$ tasks with $4$-way $8$-shot and $16$-support classification. We replicate each experiment over $10$ random seeds.}
\label{tab:metaval_add}
\end{center}
\end{table}

\begin{small}
\begin{table*}[t]
\begin{adjustbox}{width=\textwidth} 
\centering
\begin{tabular}{cccccc}
Backbone & Methods & Foraminal & \begin{tabular}[c]{@{}l@{}} Knee \\ (Acute Tear vs Not) \end{tabular} & \begin{tabular}[c]{@{}l@{}} Knee \\ (Complete tear vs Not) \end{tabular} &  Lung  \\
\toprule
& Baseline & $.24$ & $.29$ & $.32$ & $.19$ \\
& Multi-Task & $.29$ & $.34$ & $.41$ & $.27$ \\
& Vanilla ProtoNet & $.75$ & $.71$ & $.66$ & $.65$\\
& Big ProtoNet & $.57$ & $.58$ & $.53$ & $.6$\\
& Leopard & $.63$ & $.72$ & $.61$ & $.41$\\
BERT-base  & ProtoNet w/ Isotropic Gaussian & $.77$ & $.72$ & $.69$ & $.68$\\
& ProtoNet w/ Isotropic Gaussian + reg & $.78$ & $.76$ & $.71$ & $.70$\\
& Variance Aware ProtoNet (ours) & $.79$ & $.78$ & $.73$ & $.72$\\
& Variance Aware ProtoNet + reg (ours)& $.81$ & $.80$ & $.76$ & $.75$\\
\midrule
& Baseline & $.28$ & $.32$ & $.40$ & $.25$ \\
& Multi-Task & $.32$ & $.35$ & $.44$  & $.29$ \\
& Vanilla ProtoNet & $.74$ & $.73$ & $.65$ & $.67$\\
& Big ProtoNet & $.58$ & $.59$ & $.55$ & $.61$\\
BERT-base w/ Adapters & ProtoNet w/ Isotropic Gaussian & $.78$ & $.71$ & $.67$ & $.69$\\
& ProtoNet w/ Isotropic Gaussian + reg & $.80$ & $.74$ & $.72$ & $.74$\\
& Variance Aware ProtoNet (ours) & $.80$ & $.74$ & $.72$ & $.74$\\
& Variance Aware ProtoNet + reg (ours)& $.82$ & $.77$ & $.77 $ & $\mathbf{.78}$\\
\midrule
& Baseline & $.31$ & $.37$ & $.42$ & $.28$ \\
& Multi-Task & $.34$ & $.45 $ & $.47$ & $.38$ \\
& Vanilla ProtoNet & $.77$ & $.72$ & $.68$ & $.66$\\
& Big ProtoNet & $.57$ & $.59$ & $.53$ & $.61$\\
& Leopard & $.74$ & $.78$ & $.77$ & $.62$\\
Clinical BERT  & ProtoNet w/ Isotropic Gaussian &$ .78$ & $.74$ & $.71$ & $.68$\\
& ProtoNet w/ Isotropic Gaussian + reg & $ .80$ & $.76$ & $.74$ & $.71$\\
& Variance Aware ProtoNet (ours) & $.82$ & $.79 $& $.76$ & $ .74$\\
& Variance Aware ProtoNet + reg (ours)& $.84$ & $.81$ & $.79$ & $.76$\\
\midrule
& Baseline & $.35$ & $.42$ & $.45$ & $.33$ \\
& Multi-Task & $.37$ & $.45$ & $.49$ & $.37$ \\
& Vanilla ProtoNet & $.76 $& $.74$ & $ .70$ & $.67$\\
& Big ProtoNet & $.58$ & $.60$ & $.57$ & $.62$\\
Clinical BERT w/ Adapters  & ProtoNet w/ Isotropic Gaussian & $.79$ & $.76$ & $.72$ & $.70$\\
& ProtoNet w/ Isotropic Gaussian + reg & $.81$ & $.77 $& $.73$ & $.72$\\
& Variance Aware ProtoNet (ours) & $ .83$ & $.81$ & $.76$ & $.73$\\
& Variance Aware ProtoNet + reg (ours)& $\mathbf{.85}$ & $\mathbf{.82}$ & $\mathbf{.81}$ & $.77$\\
\bottomrule
\end{tabular}
\end{adjustbox}
\caption{Table showing F1 scores of few shot models with BERT-base and Clinical BERT backbones in downstream classification tasks.}
\label{tab:classification_add}
\end{table*}
\end{small}
\begin{algorithm*}[h]
\caption{PyTorch style Pseudocode for Variance Aware ProtoNets}\label{alg:cap}
\DontPrintSemicolon
\tcc{f: Encoder Network}
\tcc{N: dimension of the representation}
\tcc{c: Number of classes or ways }
\tcc{k: Shots or number of examples per class in the query set}
\tcc{m: Supports per class in the support set}
\tcc{dist: Pairwise squared Euclidean distance function}
\tcc{loss\_fn: Cross-Entropy Loss function}
\tcc{$\lambda$: regularizer}
\KwInput{Sample a set $L$ of labels, mini-batch of Support set $S_L$, Query set $Q_L$ }
\tcc{Compute statistics for each class in the Support set}
sorted\_labels = torch.sort(support\_labels) \tcp*{sort the labels in the support set}
$c$ = len(s.values.unique()) \tcp*{Number of ways}
support\_sorted = support[sort.indices] \\
labels\_sorted = labels[sort.indices] \\
embeddings\_support = $f$(support\_sorted) \tcp*{$m*c \times N$}
$m$ = embeddings\_support.shape[$0$]//$c$ \tcp*{support per class}
embeddings\_support = embeddings\_support.reshape($c, m, -1$) \tcp*{$c \times m \times N$}
 support\_mean = embeddings\_support.mean(dim=$1$) \tcp*{$c \times N$}
{\color{teal}
support\_var = torch.var(embeddings\_support, dim=$1$)**$2$} \tcp*{$c \times N$}

\tcc{Get embeddings for the query set and compute distances from the support}
query = $f(Q_L)$ \tcp*{$k \times N$} 
logits = dist(query, support\_mean) + {\color{teal} torch.sum(support\_var,dim=$1$))} \tcp*{$k \times c$, adding trace to the distance matrix}
loss = loss\_fn(-logits, query\_labels) + {\color{teal}$\lambda$ * (torch.norm(support\_var, dim=$1$))/c} \tcp{Regularizer term}

\end{algorithm*}

\section{Hyperparameters and Additional Experimental Details}~\label{sec:hparam}
In this section, we will describe the hyperparameters used for experiments on our internal and public datasets and explain some of the design choices. Table~\ref{table: Hyperparameters} shows the best hyperparameters used for our experiments. For our internal dataset, we use the Pfeiffer configuration in the adapter implementation from~\citep{pfeiffer-etal-2020-adapterhub}, whereas for the public datasets we use the exact implementation and configuration as in~\citep{wang2021gradtask} for a fair comparison to the results reported there. For all vanilla ProtoNet experiments, we use the Euclidean distance as it outperforms the cosine distance. All BERT models without adapters are trained with $8$ shots and $8$ support due to memory considerations. We choose learning rate and the variance regularizer for each model from
$\{1e-5, 2e-5, 5e-5, 1e-4\}$ and $\{1e-4, 1e-3, .01, .1, .5 \}$ based on the validation performance. For all the experiments, a dropout layer is added after the final BERT layer. 

For our internal dataset, we also experiment with $\{2,3,4\}$ ways and $\{4,6,8\}$ shots and $\{4,6,8,12,16\}$ support. The experiments with $2$-way and $3$-way produce poor results on our downstream tasks irrespective of the number of shots and support. During training with $4$-way, the meta-validation results for lower support show worse performance than the numbers reported in table~\ref{tab:results0}. We believe that it is caused by the high variability between the various groups of samples of a given class. Finally our downstream performance is best for models that are trained on higher number of shots. 

In case of ProtoNets, there is no adaptation during testing. The validation set is used to compute prototypes to query the test set. However, in case of Leopard, there is an additional few shot adaptation step that occurs as outlined in~\citep{wang2021gradtask}. In this case, the validation set is used for the adaptation and also as the support set for querying the test set.

\begin{table}[h]
\begin{center}
\resizebox{\columnwidth}{!}
{
\begin{tabular}{l l l }

Hyperparameter Type &  Internal Dataset & Public Datasets \\
\toprule
Epochs & $30$ & $40$ \\
Sequence Length & $128$ & $128$ \\
Optimizer &  AdamW & AdamW  \\
Learning Rate & $3e-5$ & $2e-5$ \\
Weight Decay & $1e-4$ & $1e-4$ \\
Gradient Clip & $3$ & $2$ \\
Early Stopping & Yes & Yes \\
Learning Rate Scheduler & Linear & Linear \\
Dropout & $.1$ & $.1$ \\
Shots & $8$ & $8$ \\
Number of supports & $16$ & $16$ \\
Variance Regularizer & $.1$ & $.01$ \\
\bottomrule

\end{tabular}
}
\caption{Hyperparameters used for all our Variance Aware ProtoNet experiments with BERT+Adapter backbones}
\label{table: Hyperparameters}
\end{center}
\end{table}

\subsection{Effect of Regularization on means and variances}~\label{sec:var_par} 
Table~\ref{tab:reg_means_var} illustrates the benefits of adding the regularization term. The regularization term not only aids in lowering the variances but also manages to push the centroids away further which we believe sheds some light on our method's success in downstream classification tasks.
\begin{table*}[h]
\resizebox{\textwidth}{!}
{
\begin{tabular}{@{}l|lll|lll@{}}
\toprule
Experiment type          & Without regularization &      &      & With regularization &      &      \\ \midrule
Task Names & \begin{tabular}[c]{@{}c@{}} Distance \\ between centroids \end{tabular} & \begin{tabular}[c]{@{}c@{}} Norm of \\
class $0$ variance \end{tabular} & \begin{tabular}[c]{@{}c@{}} Norm of \\
class $1$ variance \end{tabular} & \begin{tabular}[c]{@{}c@{}} Distance \\ between centroids \end{tabular} & \begin{tabular}[c]{@{}c@{}}Norm of \\ class $0$ variance \end{tabular} & \begin{tabular}[c]{@{}c@{}} Norm of \\ class $1$ variance \end{tabular} \\
Lung                     & $2.97$                   & $1.12$ & $1.31$ & $3.96$                & $1.15$ & $1.08$ \\
Foraminal                & $3.65$                   & $1.71$ & $1.59$ & $4.17$                & $1.62$ & $1.38 $\\
Knee (ACL complete tear) & $4.12$                   & $1.97$ & $2.10$ & $5.01$                & $1.87$ & $1.85$ \\
Knee (ACL acute tear)    & $3.95$                   & $1.53$ & $1.49$ & $4.32$                & $1.27$ & $1.35$ \\
\bottomrule
\end{tabular}
}
\caption{Table showing the showing the class statistics with and without regularization. Higher Distance and Lower variance is better.}
\label{tab:reg_means_var}
\end{table*}
We also carry out various ablation studies by changing the regularization hyperparameter. 
\begin{figure}
    \centering
    \includegraphics[width=\columnwidth]{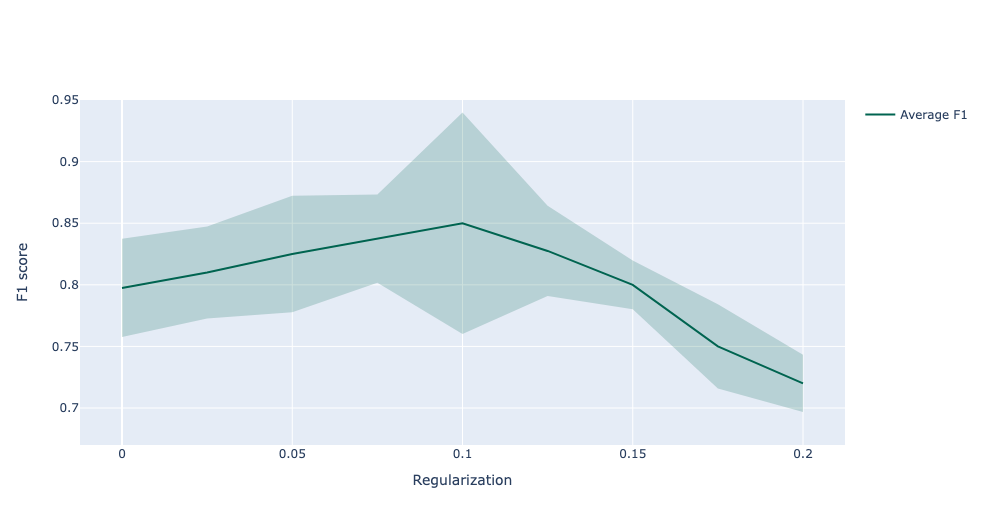}
    \includegraphics[width=\columnwidth]{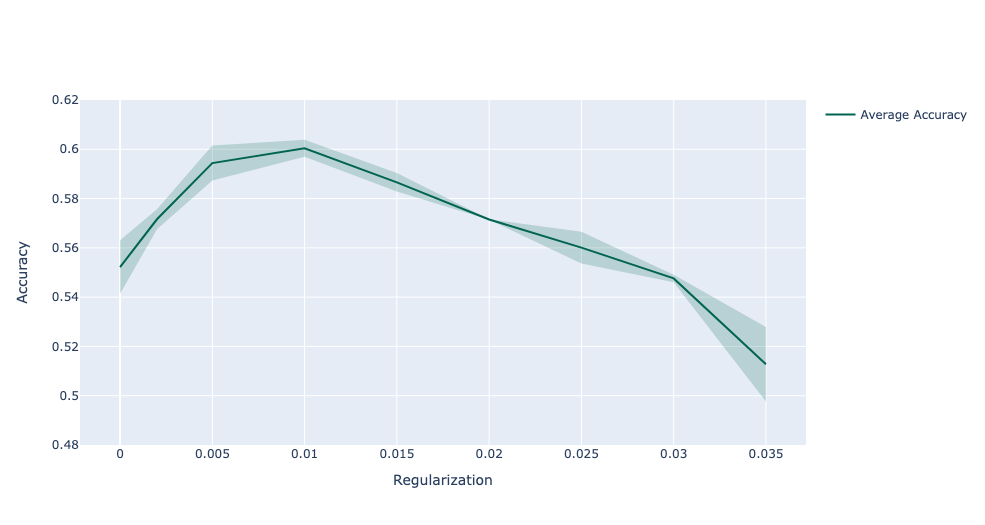}
    \caption{Figure showing the effect of changing the regularization hyperparameter. Top: Figure showing the F1 score averaged over our $4$ internal datasets. Bottom: Figure showing $16$-shot accuracy averaged over $13$ public datasets. We see a similar trend for $4$ and $8$-shot accuracies for these public datasets as well. }
    \label{fig:reg_acc}
\end{figure}
\subsection{Metric and other modeling choices}~\label{sec:wass_mot}
In~\citep{protonets}, only the sample means (i.e. means of the support vectors) are used to estimate the true population mean. In fact, by the Central limit theorem, we can use the sample variance (after normalization) to get an unbiased estimate of the population variance. Unlike the original work, we sought to use this extra information to better understand the class distribution. The Gaussian assumption is strong but it is motivated by the fact that it allows us to compute Wasserstein distances in a computationally tractable manner. Finally to motivate the choice of using the Wasserstein distance instead of a Bergman divergence like KL divergence, consider the following motivating example, $N_1(\mu_1, \Sigma_1),  N_2(\mu_2, \Sigma_2)$ be $2$ Gaussians and for simplicity assume: $\Sigma_1 = \Sigma_2 = wI$ and $\mu_1 \neq \mu_2$. With these assumptions, $W^2 = ||\mu_1 -  \mu_2 ||^2 $ And $D_{KL} = \frac{1}{2w}|| \mu_1-\mu_2 ||^2$. Note that Wasserstein distance does not change if the variance changes ($w$ can be arbitrarily large) whereas the KL divergence does. In fact, this is pointed out in~\citep{ding2022fewshot} where their goal is to create spherical Gaussians with large radii. However, we found that having large variance produces worse results in our downstream tasks. Finally similar dependence on variance is in play if one computes a simple likelihood of the sample in the class distributions.

\section{Public Benchmarks}~\label{sec:benchmark}
In this section, we describe the training procedure for the public benchmark datasets. The baseline results are taken from~\citep{wang2021gradtask, bansal-etal-2020-learning}. We have followed the same meta-training procedure as described in~\citep{wang2021gradtask}.  Specifically, for meta-training,  WNLI (m/mm), SST-$2$, QQP, RTE,
MRPC, QNLI, and the SNLI datasets~\citep{bowman-etal-2015-large} are used. The
validation set of each dataset is used for hyperparameter searching and model selection. The  models are trained by sampling episodes from the meta-training tasks. The sampling process first selects a dataset and then randomly selects $m$ examples for each class as
the support set and another $k$-shots as the query set and the probability of a selected task is proportional to the square root of its dataset size~\citep{bansal2020self}. For meta-testing, we use $13$ datasets ranging from NLI, text classification and sentiment analysis. For the models and datasets marked with $*$, we use the results reported in~\citep{bansal-etal-2020-learning} and for those datasets, we use the code from~\citep{wang2021gradtask} to generate the results for ProtoNet with Bottleneck Adapters while the rest of the results are taken from~\citep{wang2021gradtask}. We reuse their implementation and configuration of their adapters but modify the loss function with the Wasserstein distance along with our variance regularization term. Table~\ref{tab:pubres} shows the superior performance of our method beating all the baselines. For detailed hyperparameters, please see section~\ref{sec:hparam}. Our method without the variance regularization term shows similar performance to that of the Leopard baselines. For the isotropic variant method, it shows similar performance to Leopard with the variance regularization term and worse without. 

\section{Stability of the Prototypes}~\label{sec:stability}
\begin{figure}[h]
    \centering
    \includegraphics[width=\columnwidth]{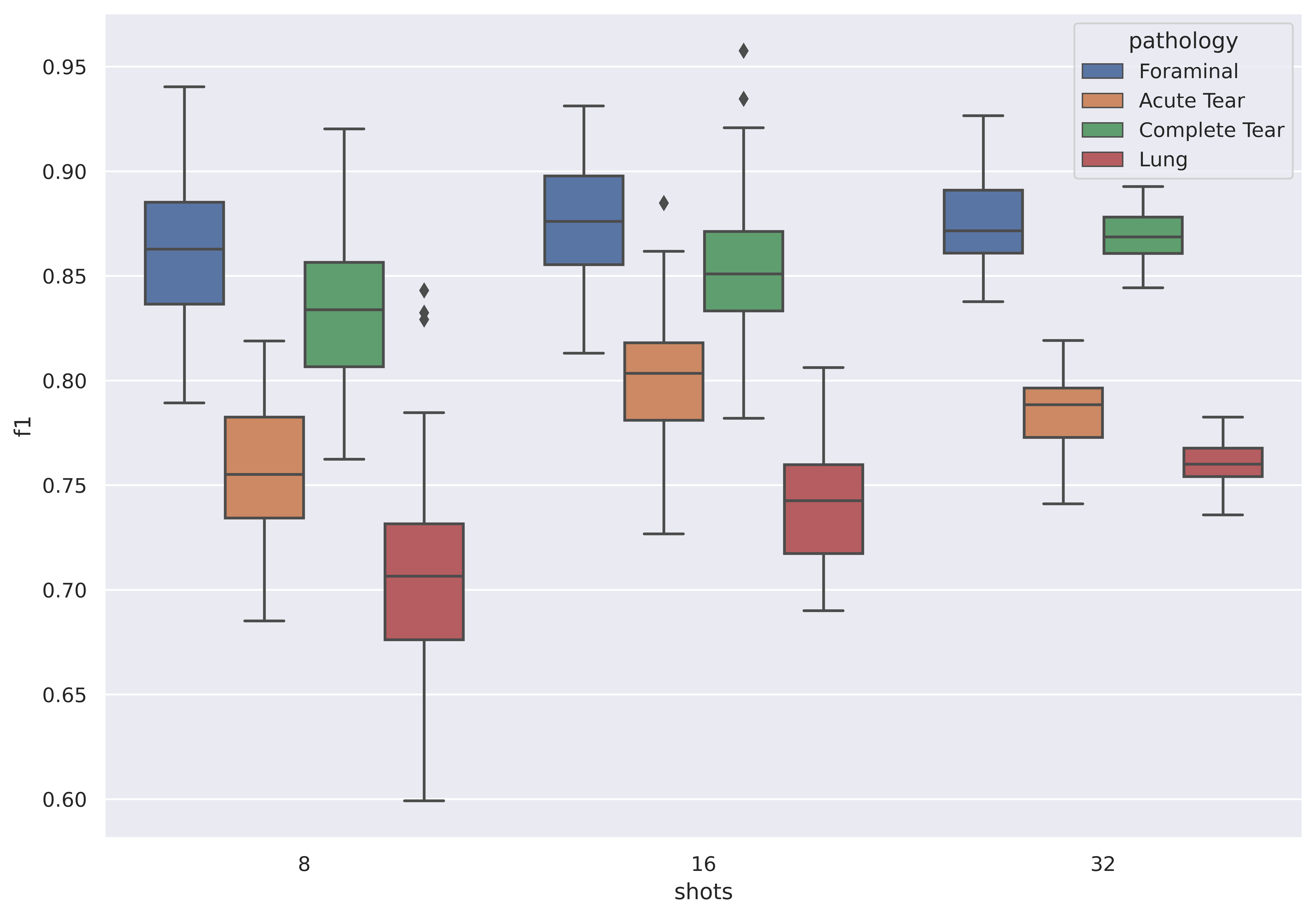}
    \caption{Figure showing stability of the prototypes. We sample $k$ examples $50$ times to construct the prototypes and the standard deviations. }
    \label{fig:stability}
\end{figure}

For simplicity, we use our entire validation sets to compute prototypes. In this section we show how our results vary if we choose a subset of our validation set to create the prototypes. The figure~\ref{fig:stability} shows the F1 scores when a subset of the data is used to compute the prototypes and the variances for a given class. 
\begin{figure}[h]
    \centering
    \includegraphics[width=\columnwidth]{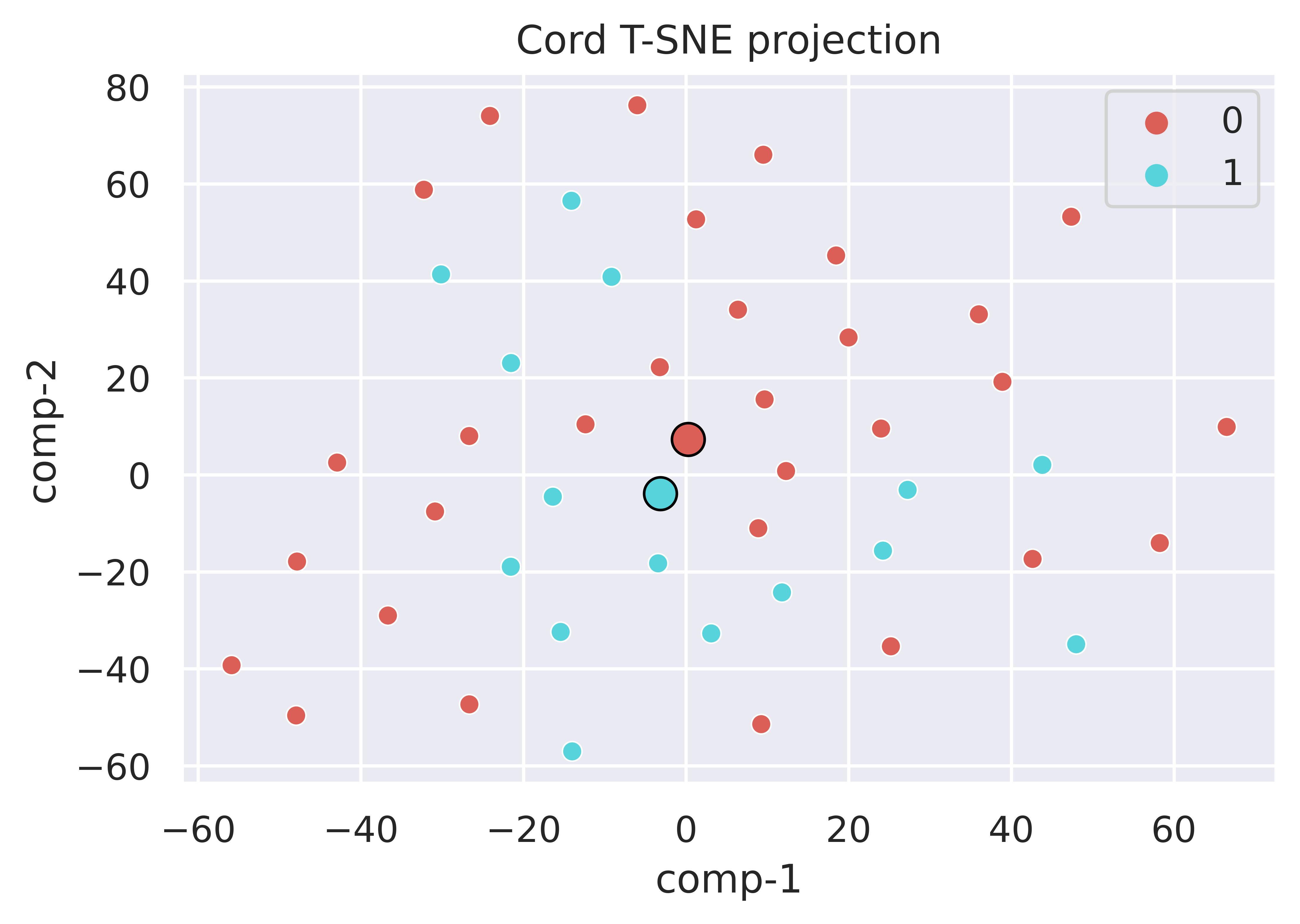}
    \includegraphics[width=\columnwidth]{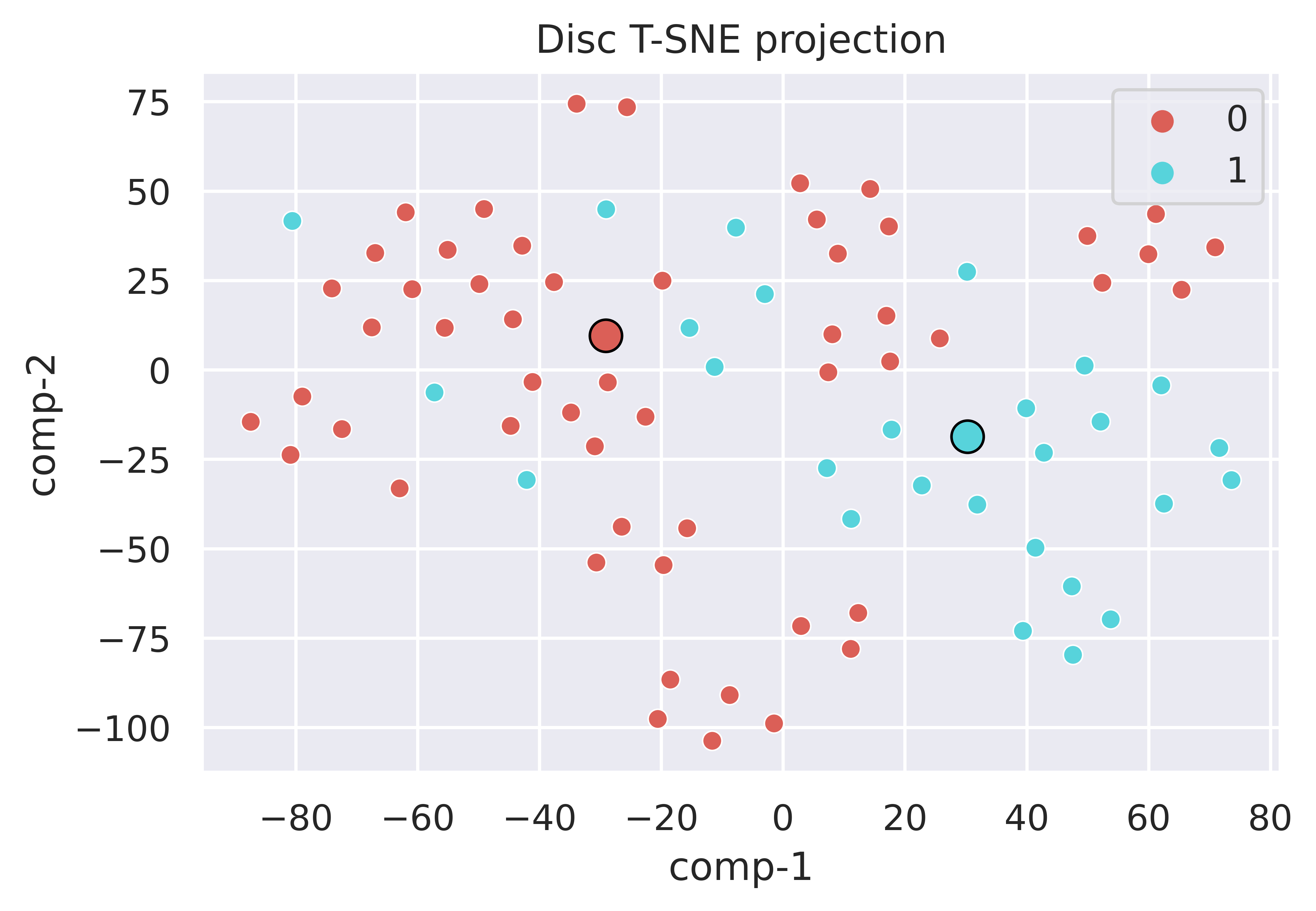}
    \caption{T-Sne projections of our Cord and Disc Data. The prototypes for cord classes are very close while the prototypes for disc are well separated. However the large variance in the disc classes causes bad performance. }
    \label{fig:failures}
\end{figure}
\section{Failure Cases}~\label{sec:fails}
We also test our models on few additional tasks like \textbf{(i)} predicting the severity of disc herniation in our cervical dataset and \textbf{(ii)} predict the presence of cord compression at various motion segments in our internal dataset on the lumbar spine. Our models achieve an F1 score of $.51$ and $.39$ respectively. The figure~\ref{fig:failures} shows how the classes are distributed. We attribute the failures to the poor separability between classes and the high variance in the data distribution.

It is an ongoing project to understand what makes our model work for these downstream tasks and why our model works on some tasks and not others. We hope that by simply increasing the diversity of our training data or applying newer adapter architectures like Mix-and-Match Adapter~\citep{mam} and Compacter~\citep{compacter}, our current methods will work on a wide range of downstream pathologies.
\end{document}